\documentclass[letterpaper, 10 pt, conference]{ieeeconf}
\IEEEoverridecommandlockouts
\usepackage{amssymb,amsmath}
\usepackage{algorithm}
\usepackage{booktabs}
\usepackage{multirow}
\usepackage{adjustbox}

\usepackage{amsmath,amssymb,amsfonts}
\usepackage{algorithm}
\usepackage{algpseudocode}
\usepackage{graphicx}
\usepackage{textcomp}
\usepackage{hyperref}
\usepackage{xcolor}
\usepackage{multirow}

\def\BibTeX{{\rm B\kern-.05em{\sc i\kern-.025em b}\kern-.08em
    T\kern-.1667em\lower.7ex\hbox{E}\kern-.125emX}}

\begin{document}

\makeatletter
\newcommand{\linebreakand}{%
  \end{@IEEEauthorhalign}
  \hfill\mbox{}\par
  \mbox{}\hfill\begin{@IEEEauthorhalign}
}
\makeatother

\title{\LARGE \bf From the Desks of ROS Maintainers: A Survey of Modern \& Capable Mobile Robotics Algorithms in the Robot Operating System 2}

\author{
\authorblockN{Steve Macenski}
\authorblockA{\textit{ROS~2 Project Lead} \\
\textit{Samsung Research}\\
s.macenski@samsung.com}
\and
\authorblockN{Tom Moore}
\authorblockA{\textit{Robot Localization Maintainer} \\ 
\textit{Locus Robotics}\\
tmoore@locusrobotics.com}
\and
\authorblockN{David V. Lu}
\authorblockA{\textit{ROS~1 Co-Maintainer} \\
\textit{Metro Robots}\\
davidvlu@gmail.com}
\and
\linebreakand 
\authorblockN{Alexey Merzlyakov}
\authorblockA{\textit{ROS~2 Developer} \\ 
\textit{Samsung Research}\\
alexey.merzlyakov@samsung.com}
\and
\authorblockN{Michael Ferguson}
\authorblockA{\textit{ROS~1 Co-Maintainer} \\
\textit{Cobalt Robotics}\\
fergs@cobaltrobotics.com}
}


\maketitle
\begin{abstract}
The Robot Operating System~2 (ROS~2) is rapidly impacting the intelligent machines sector - on space missions, large agriculture equipment, multi-robot fleets, and more.
Its success derives from its focused design and improved capabilities targeting product-grade and modern robotic systems.
Following ROS~2's example, the mobile robotics ecosystem has been fully redesigned based on the transformed needs of modern robots and is experiencing active development not seen since its inception.

This paper comes from the desks of the key ROS Navigation maintainers to review and analyze the state of the art of robotics navigation in ROS~2. 
This includes new systems without parallel in ROS~1 or other similar mobile robotics frameworks.
We discuss current research products and historically robust methods that provide differing behaviors and support for most every robot type.
This survey consists of overviews, comparisons, and expert insights organized by the fundamental problems in the field.
Some of these implementations have yet to be described in literature and many have not been benchmarked relative to others.
We end by providing a glimpse into the future of the ROS~2 mobile robotics ecosystem.

\end{abstract}


\section{Introduction}
\label{sec:introduction}

The Robot Operating System 2 (ROS~2) was redesigned from the ground up to meet the requirements of modern robotics research and commercial products in warehouses, urban settings, agriculture, aerospace, autonomous driving, and more.
It also provides previously missing capabilities such as security, multi-robot communications, and embedded/real-time support.
Philosophically, ROS~2 was created as a production-grade robotics framework to serve the growing and maturing robotics industry \cite{ros2}.

The transition to ROS~2 has sparked a renaissance of new development in long-dormant project ecosystems.
This new development is holistic, comprising algorithmic refreshes, new features, documentation, and increased maturity inspired by ROS~2's philosophical and procedural changes.
The mobile robotics ecosystem in ROS~2 has seized this opportunity to rethink and redesign based on the needs of modern systems.

\textit{Nav2} is the major project within the mobile robotics ecosystem and provides a next-generation autonomous mobile robotics navigation framework and system \cite{nav2}.
The original Navigation Stack was developed at Willow Garage for the PR2 robot and based around an unconfigurable state machine, {\tt move\_base} \cite{nav}.
It largely supported circular differential-drive and omnidirectional research platforms.

\textit{Nav2} was built from the ground up to leverage innovations in behavior trees and contains refreshed algorithms.
It supports all common forms of classical robots (e.g. differential, omnidirectional) as well as newer robot platforms such as Ackermann, large, non-circular, and quadruped robots.
It maintains production-grade quality standards with 90\% unit test coverage and seeks to re-centralize development across academia and industry.

The \textit{Nav2} project and ROS~2's mobile robotics ecosystem have been developing new capabilities at a pace not seen since the days of Willow Garage. 
New algorithms, features, and behavior trees are being added on a quarterly basis as the result of open collaborations between companies, individuals, and academia.
Over the past two years, a compelling set of new algorithms and behaviors have been made available - some of which have no comparable capability in other similar frameworks (ROS~1, MRPT, etc) - and have yet to be compared in literature \cite{mrpt}.
The introduction of Nav2 (\cite{nav2}) showcases many of these new algorithms and behaviors working on commercial robots in highly dynamic environments navigating over 2 marathons in length on an active college campus. 
The authors recommend reviewing this document for a practical and reproducible example of the composed capabilities described in this paper.  

This survey comes from the desks of core ROS developers and maintainers to be the authoritative discussion on the state-of-the-art in ROS~2 robotics navigation.
This survey includes overviews, analysis, and unique expert insight into the algorithms and capabilities, new and old.
We discuss and compare current research products with historically robust methods.
Brief benchmarking within provides metrics for the first time for the new key methods.
Finally, this survey ends with a glimpse into the future of ROS~2 mobile robotics capabilities. 

\section{Overview of Global Path Planners}
\label{sec:planners}

Global path planning seeks to address the foundational problem of finding optimal sequences of valid configurations to determine a route through an environment. 
A path may have different levels of fidelity. It may comprise sequences of neighboring cells in a grid map, feasible and continuous sequences, or even vehicle dynamics.
For global path planning in Nav2, only up to kinematic constraints are considered.
The hybrid planning schema utilized establishes that the global planner's objective is to find a route through the environment, while a local trajectory planner follows this route considering system dynamics and other criteria \cite{hybrid}. 

The definition of optimal in global path planning can vary with application, and at its core, the algorithms are all extensions of Dijkstra's algorithm in trying to find the path with minimal cost \cite{dijkstra1959note}. 
In Nav2, the cost is largely determined by the \textit{cost map} which is discussed at length in Sec. \ref{sec:costmap}. 
In practice, the result is usually the shortest path that does not navigate too closely to any obstacle, as seen in Figure \ref{fig:InfeasiblePlanners}.

Path planners are typically either search- or sampling-based.
Sampling-based planners, like RRT and its variants, are prevalent for higher dimensional planning problems and can provide near-optimal paths - although they commonly require post-processing before use \cite{rrt}. 
However, search-based planners in lower dimensional state spaces, like those found in mobile robotics, are typically faster and produce truly optimal paths when heuristics are admissible and consistent.
Search-based planners also provide more predictable execution times important to dynamic replanning over complex large-scale environments.

Thus, Nav2 provides primarily search-based global planning methods.
The furnished algorithms cover the full spanning set of both circular and arbitrarily-shaped robot models, including differential-drive, holonomic, Ackermann, and legged robot platforms.
Namely, these are Navigation Function, 2D-A*, Theta*, Hybrid-A*, and State Lattice planners. 
All of the planners are cost-aware, meaning they consider the range of cost map risk values populated from sensor data, perception pipelines, or semantic information such as maps - not solely binary obstacles.

For comparison, a AMD Ryzen 5 5600X CPU with Ubuntu 20.04 was utilized to benchmark each planning algorithm.
The experiment contained 1,000 randomly generated start and goal poses through a 20\% randomly occupied 10,000$m^2$ map.
Each start and goal pairing was made to be at least 3 meters apart and known to be reachable.
All of the algorithms are analyzed with the same 1,000 paths and uses the provided default settings.
The results are shown in Table \ref{tab:planners}.

\begin{table*}[!ht]
 \caption{Path Planners Comparison, Bolded Within 3\% of Best}
 \label{tab:planners}
 \begin{center}
 \begin{tabular}{ |c|c|c|c|c|c|c|c|c|c|c|c|c|c|c| }
   \hline
   Path Planner & Plan Time (ms) & Path Length (m) & Feasible & Circular & Non-Circ. & Diff. & Omni. & Ackermann & Legged \\
   \hline
   \hline
   NavFn & 61.02 & 52.25 & - & \checkmark & - & \checkmark & \checkmark & - & - \\
   \hline
   Lazy Theta*-P$^1$ & 94.42 & \textbf{50.28} & - &  \checkmark & - & \checkmark & \checkmark & - & - \\
   \hline
   Smac 2D-A*$^1$  & 88.82 & \textbf{49.65} & - &  \checkmark & - & \checkmark & \checkmark & - & - \\
   \hline
   Smac Hybrid-A*$^1$ & \textbf{38.77} & \textbf{50.78} & \checkmark &  \checkmark & \checkmark & \checkmark & - & \checkmark & \checkmark \\
   \hline
   Smac State Lattice$^1$ & \textbf{39.40} & \textbf{50.51} & \checkmark &  \checkmark & \checkmark & \checkmark & \checkmark & \checkmark & \checkmark \\
   \hline
 \end{tabular}
 \end{center}
 \hspace{70pt} $^1$ Exclusive to ROS~2. \hspace{50pt} An inter-modal algorithm selection summary can be found in Appendix \ref{algo_selection}.
\end{table*}

\subsection{Holonomic Planners}
\label{sec:planners_infeasible}

Nav2 provides a variety of holonomic path planners.
This class of planner is most useful for omni-directional or differential-drive robots, particularly those that may be approximated as circular.
Non-circular robots that are small relative to their environment may be treated as circular for global planning purposes if there is no navigable corridor too narrow to accommodate an in-place rotation.
While differential-drive robots cannot move at any angle, they may perform in-place rotations to a given requested heading if required.
Thus, many forms of differential-drive platforms may effectively use holonomic planners.

These planners generally create paths via traditional grid-based algorithms.
Since they do not track robot orientation during grid-search, these algorithms collision check using point-costs of the visited cells in an inflated or convolved cost grid, which are discussed further in Sec. \ref{sec:costmap}.
Those included in Nav2 are Navigation Function (commonly known as \textit{NavFn}), Theta*, and smoothed 2D-A* \cite{navigation_function, theta_star, astar}.

\begin{figure}[ht]
    \centering
    \includegraphics[width=0.48\textwidth]{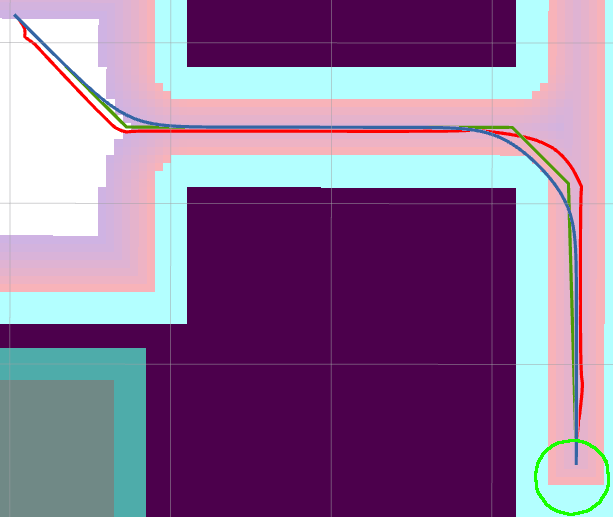}
    \caption{Example plans from the holonomic planners. NavFn in red, Theta* in green, 2D-A* in blue.}
    \label{fig:InfeasiblePlanners}
\end{figure}

\textbf{Navigation Function} planner is a modified Global Dynamic Window Approach planner based on the NF1 function \cite{navigation_function}.
It uses Dijkstra's Algorithm to expand a potential field from the goal cell to the start using a Von Neumann neighborhood.
The potential field is calculated using a quadratic kernel function rather than distances to the goal (NF1) to create a smoother potential field, and thus smoother output paths \cite{e_star}.
The kernel at a cell is found based on the cell's cost and the value of its neighboring cells' potentials.
The planning costs at each cell are linearly remapping from the cost grid to produce a minimum cost to travel, or neutral cost.
Therefore, each cell's cost ranges from the neutral to maximum cost when populating the potential field. 
This generates increasing costs each wavefront until the start cell is found.
Finally, to traceback the path, gradient descent is applied from the start cell in the potential field in half resolution sized increments to produce a uniformly distributed path.

This planner is the oldest algorithm in Nav2 - derived from the ROS~(1) Navigation Stack and created in the first public commit to the project.
Regardless of age, it is highly optimized with managed priority buffers and remains as one of the most effective holonomic planners available, typically taking between 15 - 175 ms.
It has been proven by dozens, if not hundreds, of robotics companies and researchers over the past ten years in a variety of industry, consumer, and service applications. 
Due to its speed and long-term stability, the Navigation Function planner is a good general-purpose option and remains the default planning algorithm in Nav2. 

Although it is the fastest holonomic planner, it produces paths 5\% longer by design, Table \ref{tab:planners}.
The neutral cost balances cell costs with path lengths when generating the potential field.
It was carefully selected empirically to produce high-quality paths in a variety of environments, including routing down centers of hallways, maintaining safe margins from obstacles when space is allotted, and lacking in excessive movements. 
The planner is made to prefer slightly longer routes to minimize cost, which provides additional margins from collision, which can be seen on Fig. \ref{fig:InfeasiblePlanners}.
However, the neutral cost may be tuned for different behaviors.

\textbf{Lazy Theta*-P} is an any-angle path planner based on A* \cite{theta_star}.
The base Theta* technique modifies A* with line-of-sight (LoS) checks between the parent of a node and the neighbor being expanded from that node.
If there is LoS, that connection is considered to short-cut the current visited node.
In this implementation, each grid cell is a potential search node.
During planning, the visited nodes may be repeatedly short-cutted until there is no longer a valid line-of-sight - resulting in the possibility of any-angle LoS between potentially distant parent and child nodes.

Line-of-sight checks are implemented using Bresenham's ray-casting algorithm \cite{bresenham}.
The LoS check evaluates if there is a collision with the environment between the two nodes; but the Theta* planner in Nav2 also evaluates if the integrated cost of the line is below a configurable threshold to be considered valid.
This process breaks LoS for a given plan heading to be more responsive to the localized environment in high-cost (e.g. confined) areas.
In practice, this makes the planner cost-aware, similar to other planners discussed, such that a path doesn't route unacceptably close to obstacles.
This threshold may be tuned for an optimal trade-off between segment lengths and proximity to obstacles.
After planning, the path is interpolated over its constituent LoS segments to create a regularized path. 

Nav2's Lazy Theta*-P uses a Moore neighborhood and L2 distance heuristic. 
Lazy Theta* optimizes the algorithm by reducing the number of line-of-sight checks by delaying them until they are strictly necessary \cite{lazy_theta}.
This lazy evaluation conducts only one LoS check when visiting a cell rather than for each expanded cell, yielding a reduction in expensive LoS checks.
Lazy Theta*-P builds on this by evaluating LoS on the current node's grandparent rather than parent in the graph.
Lazy Theta*-P also has the important property that every expanded cell will have definite line-of-sight with its set parent.
The speculative assumption in Lazy Theta* that a line-of-sight between the node's neighbor and parent exists is not always true.
Even if this is corrected when visited, it does not maintain the same properties as A*.
However, as a consequence, Lazy Theta*-P performs more LoS checks than Lazy Theta*, as it needs to expand more cells.


Since Theta* and its variants may produce plans at any angle, it is particularly useful for maps at non-axially aligned angles where planners restricted to 4- or 8-connected neighborhoods would produce jagged paths.
While it is common to rotate maps such that they are axially aligned with important features, that is not always practical nor possible. 
It is also useful in spaces with many long, straight corridors commonly found in office buildings, warehouses, retail stores, and similar.
Theta*'s straight line characteristics are powerful for predictability of robot behavior to bystanders, as well as providing more fine-tuned and localized alignment to the environment, Fig. \ref{fig:InfeasiblePlanners}. 

In experiments, this algorithm had the shortest path lengths, yet the longest run-time due to the expensive LoS checks.
Yet it still displays appropriate run-time performance under 100 ms making it applicable in many practical applications, Table \ref{tab:planners}.

\textbf{2D-A*} planner is the simplest of the holonomic approaches, as a cost grid-A* planner \cite{astar}.
It is implemented within the \textit{Smac Planner} Framework complimentary to the feasible planners discussed in Sec. \ref{sec:planners_feasible}.
See Appendix \ref{smac_appendix} for more information about the Smac Planner Framework.
The 2D-A* planner uses Moore neighborhoods with an L2 distance heuristic.
The travel cost between cells is derivative of the cost-aware obstacle heuristic discussed later in detail, and takes the form:

\begin{align}
    cost_{travel} = d \hspace{5pt} (1.0 + \omega_{cost} \frac{cost(x,y)}{cost_{max}})
\end{align}

where $d$ is the distance between cells, $\omega_{cost}$ is the configurable weight placed on penalizing cost, $cost(x,y)$ is the cost at a cell $(x,y)$, and $cost_{max}$ is the maximum possible cost to normalize the term relative to the penalty weight.
When the cost is $0$, it is simply the distance between cells ($1.0$ or $\sqrt{2}$), thereby the heuristic is admissible.

To smooth out the path generated from backtracing cells, the simple smoothing algorithm from Eq. \ref{simple} is employed to remove jagged edges, smooth turns, and eliminate other localized defects.
This planner uniquely enables users to downsample cost grids into different resolutions for holonomic planning to increase speed $O(N^2)$ in large spaces\footnote{Downsampled grids of half resolution results in planning times competitive with the other Smac Planner implementations.}.
Unsurprisingly, since the \textit{Smac Planners} share much of the search and penalty structures, their paths are within 3\% of each other (Table \ref{tab:planners}), where 2D-A* is the shortest largely attributable to the lack of non-holonomic constraints. 

The 2D-A* planner is most useful when being deployed in heterogeneous fleets of robots, whereas the non-holonomic or large robots are using other planners within the Smac Planner framework.
This enables all robots in a fleet to have consistent planning behavior based on analogous penalty functions trading off grid map cost and path length using the same weights.

\subsection{Kinematically Feasible Planners}
\label{sec:planners_feasible}

In contrast to the holonomic planners, Nav2 also provides a set of kinematically feasible planners - or planners that consider kinematic constraints.
These planners enable integration with modern robots such as Ackermann and legged robots, as well as providing accurate modeling of non-circular differential-drive and holonomic bases.
There are many cases where classical robots are large relative to their environment and require explicit modeling to create feasible global paths.
Since orientations are being considered, collision checking and planning may use the robot shape to generate paths that are high fidelity even through narrow spaces.
The feasible planners in Nav2 are Hybrid-A* and State Lattice, both members of the \textit{Smac Planner} framework.

\begin{figure}[ht]
    \centering
    \includegraphics[width=0.48\textwidth]{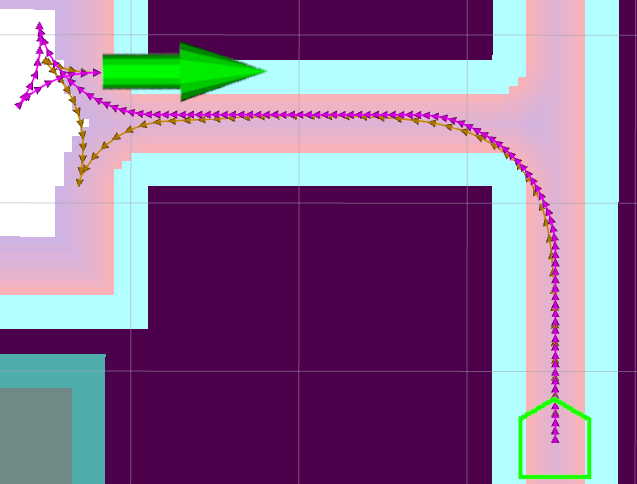}
    \caption{Example plans for kinematically feasible planners with reversing. Hybrid-A* in brown, State Lattice in magenta.}
    \label{fig:FeasiblePlanners}
\end{figure}

\textbf{Hybrid-A*} was proposed by Stanford during the DARPA Urban Challenge as an extension of traditional grid-A*~\cite{dolgov2008practical}.
It uses motion primitives to search a grid map in a tree-fashion using continuous coordinates associated at discrete cells ($x$, $y$, $\theta$).
This ensures continuous and feasible paths.
The primitives are composed of straight segments and circular arcs (left and right) of the minimal turning radius for a car-like base.
Periodic analytic expansions are applied to find exact and optimal solutions from a currently expanded pose to the goal pose. 
If this expansion is valid and collision free, it is included in the route to complete the planning process.
The search primitives and analytic expansions can be computed utilizing using Dubin or Reeds-Shepp models, depending if reversing is a permissible action.

Thus, this method is specifically designed for car-like or Ackermann vehicles; however, it has utility for legged and large robots beyond naked feasibility of non-circular platforms.
Limiting path curvature can be of practical importance for high-speed robot systems to prevent centripetal force from flipping over legged platforms or dumping unsecured loads off of differential or holonomic robots (e.g., material handling, delivery).

The Hybrid-A* planner provided by Nav2 within the \textit{Smac Planner} framework provides a few additional unique variations.
Rather than using solely binary collision information, all values in the cost grid are utilized to steer the feasible search via an admissible cost-aware obstacle heuristic function.
This leads optimal paths into lower-costed regions away from obstacles, respects user generated cost grid constraints (such as keepout zones or penalizing narrow spaces), and improves path quality.
This is sufficiently effective that expensive post-processing of the paths proposed in \cite{dolgov2008practical} is not strictly necessary to improve paths' smoothness and safety margins from obstacles.
Moreover, because expanded search nodes are generally farther from obstacles, optimizations in collision checking are made which further improve run-time performance.

Additionally, limits are placed on the maximum length of analytic expansions to prevent excessive bypassing of heuristic search.
Limiting the expansion distance prevents suboptimal behaviors that would have been penalized in the search and/or cost grid, such as reversing for long distances or entering sensitive high-cost zones.
The role of analytic expansions are to find a path to the exact goal, thus, it is most applicable in close proximity to the goal pose. 
Reducing the expansion length impacts performance negligibly. 
A typical run-time for this planner is between 10-250 ms in practical large deployed environments.

\textbf{State Lattice} was utilized by Carnegie Mellon University in the DARPA Urban Challenge \cite{CMUDARPA,pivtoraiko2007optimal}.
Like Hybrid-A*, it uses continuous coordinates associated at discrete cells to create feasible paths from motion primitives.
However, it leverages offline computed motion primitives that establish a state lattice graph rather than expanding using trees.
It expands A* as a structured lattice graph and connects primitives at precise bin quantizations rather than at arbitrary locations.
The minimum control set, or motion primitives for State Lattice, may be computed based on an arbitrary motion model, not only for Ackermann vehicles or restrictions based on minimum curvature. 
This can handle novel or atypical drive-train constraints, such as those found on planetary exploration rovers, legged robots; or differential-drive in-place rotations and omnidirectional lateral motions \cite{mars}. 
Thus, it is a generalized class of planner for any robot drive-train type when matched with a complimentary minimum control set - making it a powerful addition to Nav2\footnote{The provided control set generator yields curvature minimizing trajectories with optional in-place rotations and lateral motions to support Ackermann, Legged, Diff., Omni. robots.}.

The State Lattice planner in Nav2 provides a number of noteworthy optimizations to improve performance beyond historical libraries such as SBPL, to 20-300 ms \cite{SBPL}.
Classical State Lattice planners struggle to efficiently determine the best route through complex scenes using only  distance-based heuristics.
They also labor to find paths to precise goal poses with solely search. 
The cost-aware obstacle heuristic and analytic expansions are also shared with this planner to resolve these issues, respectively.
In providing the Smac Hybrid-A* capabilities, this planner improves performance by an order of magnitude and provides remarkably similar results to Hybrid-A* both in compute time and path length/quality, show in Table \ref{tab:planners}.
The path lengths and planning times are within 0.5\% and 1.6\% of each other, respectively. 
This is a powerful result: Nav2 contains two planners with analogous behavior each covering important niches to establish support for the breadth of possible robot platform types.
Therefore, effectively any robot base can be used with Nav2 to create kinematically feasible plans, which are also akin to any other robot type's plans - forming consistency across heterogeneous fleets.
Fig. \ref{fig:FeasiblePlanners} illustrates two paths produced by Hybrid-A* and State Lattice planners.

Note that both of these planners and the 2D-A* discussed previously derive from the \textit{Smac Planner} framework.
A number of shared optimizations, pre-computations, and caching have been made within the framework to be competitive with established planning algorithm implementations in planning time, behavior and path length.
The Smac Hybrid-A* and State Lattice planners are notably faster than their 2D-A* counterpart in large part due to an initially reduced cost-aware obstacle heuristic resolution used to steer the feasible search, which is later upsampled as nodes are visited.

\section{Overview of Local Trajectory Planners}
\label{sec:trajectory}
Once a global path planner or similar has found an acceptable route through the environment, another technology needs to convert the intended route to navigate into a stream of velocity commands for the mobile platform to track.
These commands generally attempt to follow the path, but may also consider objectives such as reducing likelihood of collision, improving smoothness, minimizing important objectives (curvature, acceleration, jerk), or may implement other primitive behaviors.
This is the work of the (Local) Trajectory Planner, which generates collision-free trajectories and issues current velocity commands to the base's motor controller.
This completes the Hybrid Planning schema introduced in Sec. \ref{sec:planners}, in that global planners find an acceptable route and trajectory planners follow it, potentially considering additional system constraints.

Due to its close coupling to the mobile robot base, trajectory planners are often referred to as \textit{controllers} and the selection of the best trajectory planner will be highly dependent on the robot kinematics. For an abbreviated set of recommendations see Appendix \ref{algo_selection}.

There are a wide variety of trajectory planning techniques to satisfy different robot base types and applications.
The behavioral needs of an industrial warehouse automation robot differs as much from a hospitality robot as a circular differential-drive base differs from an Ackermann-steering car.
Thus, there are different classes of techniques such as reactive, predictive, geometric, machine-learning, and control-law derived trajectory planners.
Furthermore, trajectory planners often need specialized tuning to match the kinematic and dynamic characteristics of a particular robot base.
While techniques are portable between robots, they typically require the most involved and intensive tuning of any system on a mobile robot.

Nav2 includes several implementations of trajectory planners to meet the spanning set of needs for differential-drive, omnidirectional, legged, and Ackermann style vehicles in both circular-optimized and arbitrary-shaped forms.
Namely, the Dynamic Window Approach (DWA), Regulated Pure Pursuit (RPP), Model Predictive Path Integral (MPPI), and Rotation Shim controllers. In addition to the trajectory planners that are part of the Nav2 project, there are several trajectory planners available in the broader ROS~2 ecosystem, such as the Timed Elastic-Band (TEB) and the Graceful Controller.
Table \ref{tab:controllers} compares these controllers behaviors and key attributes.

Nav2 passes each trajectory planner a rolling-zone local cost map, planned path, and positional information to utilize along with any additional information it obtains itself through topics, parameters, or services.
New to Nav2, run-time configurable plugin interfaces were added to abstract out commonalities between many controller implementations.
These support dynamically set speed limits, navigation goal checking criteria, and progress checking algorithms.

\begin{table*}[!ht]
 \caption{Trajectory Planners Comparison}
 \label{tab:controllers}
 \begin{center}
 \begin{tabular}{ |c|c|c|c|c|c|c|c|c|c|c|c|c|c| }
   \hline
   Local Trajectory Planner & Max. Frequency (Hz) & Type & Circular & Non-Circ. & Diff. & Omni. & Ackermann & Legged \\
   \hline
   \hline
   Model Pred. Path Integral$^1$ & 125 & Predictive & \checkmark & \checkmark & \checkmark & \checkmark & \checkmark & \checkmark \\
   \hline
   Timed Elastic Band & 130 & Predictive & \checkmark & \checkmark & \checkmark & \checkmark & \checkmark & \checkmark \\
   \hline
   DWB & 250 & Reactive & \checkmark & \checkmark & \checkmark & \checkmark & - & - \\
   \hline
   Graceful & $1,800$ & Control-law & \checkmark & \checkmark & \checkmark & \checkmark$^3$ & - & - \\
   \hline
   Reg. Pure Pursuit$^1$ & $> 4,000$ & Geometric & \checkmark & \checkmark & \checkmark & \checkmark & \checkmark & \checkmark \\
   \hline
   Rotation Shim$^1$ $^2$ & $>> 10,000$ & Kinematic & \checkmark & \checkmark & \checkmark & \checkmark & - & - \\
   \hline
 \end{tabular}
 \end{center}
 \hspace{20pt} $^1$ Exclusive to ROS~2 $^2$ To be paired with other algorithms $^3$ While not taking full advantage of the omni-directional base, this control law works well on powered caster bases which are not fully holonomic\hspace{15pt} An inter-modal algorithm selection summary can be found in Appendix \ref{algo_selection}.
\end{table*}

\subsection{Reactive Controllers}

Reactive trajectory planners use knowledge of the surrounding environment to construct collision-free trajectories towards the objective. 
They do so separately from global planning to ensure obstacle avoidance which is \textit{reactive} to changes in the local environment, which may be more fine grained, up-to-date, or information rich than that used in global planning.
This knowledge may include sensory information in addition to map or occupancy information, though not definitionally.
In Nav2, all planners use live sensor information fused with a map via its local cost map  (discussed in Sec. \ref{sec:costmap}).

\begin{figure}[t]
    \centering
    \includegraphics[width=0.48\textwidth]{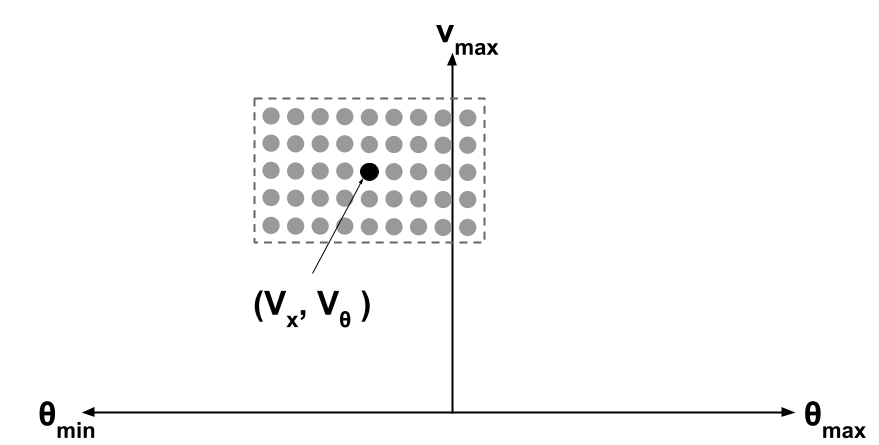}
    \caption{Visualization of the dynamic window of feasible velocity commands from the robot's current state, with 9 angular velocities ($v_\theta$) and 5 translational samples ($v_x$).}
    \label{fig:dwb}
\end{figure}

Although reactive trajectory planners often consider the environment around it, they lack full context in temporal dimensions. 
These methods typically rely on potential fields or hand-engineered heuristics which score trajectories based on the robot's current state and the localized surroundings in the current time-step only.
While this can be effective in a number of environments and boasts low computational complexity, it is often susceptible to local minima. 
Nonetheless, robust reactive control methods continue to find a great deal of application in modern robotic systems. 

In particular, the \textbf{Dynamic Window Approach} endures as a method of interest to system designers and researchers alike.
DWA is enabled by the DWB Local Planner in Nav2~\footnote{DWB as in the 'next' after DWA. It is the successor of the DWA controller found in the ROS~(1) Navigation Stack.}.

The method sub-samples a set of feasible velocity commands from the \textit{dynamic window} of achievable velocities, shown in Fig. \ref{fig:dwb}.
These are a set of candidate commands from the robot's current velocity state ($v_x$, $v_\theta$) within its dynamic limits \cite{dwa}.
This set contains constant velocities which create circular arcs when projected forward - a simple trajectory.
Each arc is scored against a set of critic functions and the best weighted average velocity is selected for a planning iteration.
The critic functions are mathematical operations which score the candidate trajectories based on various criteria optimizing for desired behavior (e.g. tracking the path and avoiding obstacles).
The original objective function takes the form in Eq. \ref{eq:dwa}:

\begin{equation} \label{eq:dwa}
\alpha \hspace{2pt} heading(v,\omega) + \beta \hspace{2pt} dist(v,\omega) + \gamma \hspace{2pt} vel(v,\omega)
\end{equation}

where $\alpha$, $\beta$, and $\gamma$ are tunable penalty weights, $heading(v,\omega)$ is a measure of distance to the goal, $dist(v,\omega)$ is the distance to the nearest obstacle on the trajectory, and $vel(v,\omega)$ encourages moving at full speed.

Within the DWB package, this method is generalized and expanded.
Rather than considering only forward translational motion, $\Dot{x}$, lateral translation is also considered for omni-directional platforms, $\Dot{y}$ (which may be set to zero for differential-drive platforms).

Further, dynamically configurable trajectory generation and critic functions may be used to expand or adapt behavior for a particular platform. 
This allows for customization of both the generation of candidate velocities and the scoring of those velocities.
Currently the following critic functions have been made available for users: goal alignment, goal distance, path alignment, path distance, prefer forward motion, anti-twirling, anti-oscillation, and traversed cost by footprint or center values.
Each contains its own relative weight and configurable parameterization to produce tunable behavior.
In addition, two unique trajectory generators are also available.
The \textit{Standard Trajectory Generator} samples trajectories based on a configurable maximum time ahead to forward simulate, while the \textit{Limited Acceleration Generator} finds trajectories based on a configurable simulation period, $\delta t$, to apply at each time-step.

DWB is currently the default local trajectory planner in Nav2, largely due to historical precedence of Base Local Planner and DWA Local Planner in ROS~(1) \cite{trajrollout}.
DWB's architecture is a refactor of these controllers from the original Navigation Stack to be more flexible with plugin-based critic functions and trajectory generators. 
The configurability of DWB is unparalleled and has been shown to be capable of any number of reactive behaviors when configured and/or customized.

However, the method suffers from numerous, highly inter-related parameters that must be tuned to achieve such performance (e.g. choice of critic functions and their relative weightings).
This process can be complex and poorly tuned DWB configurations can create extremely suboptimal behaviors. 
This tuning complexity has long been a source of critique for the ROS mobile robotics community. Many deploy un-tuned parameters and remark at poor robot behavior due to the trajectory planner, although the perception systems and global planning systems are quite mature. 
It is the maintainers' aim in the medium-term future to replace DWB as the default algorithm with MPPI - a well-tuned, modern predictive controller algorithm, discussed in Sec. \ref{sec:predictive}, once it reaches an appropriate level of maturity.

\subsection{Predictive Controllers}
\label{sec:predictive}
Predictive trajectory planners are a specialization of reactive planners which consider temporal information in the process of constructing objective-maximizing trajectories. 
They refine the previous time-step's trajectory when evaluating a current request and typically model trajectories as non-trivial sequences of velocity commands at future time intervals.
This \textit{prediction} gives the classification its distinction and makes it more robust against local minima causing a system to become in an unrecoverable or 'stuck' state.
Further, oscillation, sudden changes in velocity (and its derivatives), and other adverse behaviors are greatly reduced when using well-established methods - making a robot system exhibit more intelligent behavior.
Predictive controllers come at a cost however: they are far more computationally demanding than other approaches and are an area of evolving research.

Many predictive controllers are modelled and solved using optimization-based techniques under the umbrella of model predictive control (MPC).
MPC methods have been shown to be effective on a broad range of problems within and outside of robotics, including autonomous driving and industrial applications. 
The two predictive controllers in Nav2 both belong to the family of MPC approaches, namely the Timed Elastic Band (TEB) and Model Predictive Path Integral (MPPI) controllers - of which both are unique to the ROS mobile robotics ecosystem.


\textbf{Timed Elastic Band} (TEB) is one of the most popular trajectory planners in ROS~(1). Some global path planners generate plans that have frequent or abrupt changes in direction. Elastic bands transform such paths into smooth, deformable plans that are more appropriate for trajectory following \cite{ebands}. In the original formulation, small subsets of collision-free space, known as ``bubbles,'' are overlaid on the path and overlap in such a way as to guarantee collision-free movement between them. The path is then deformed by subjecting the bubbles to repulsive forces that represent distances from obstacles and contracting forces that eliminate excess slack in the path.

TEB is an extension of Elastic Bands that also accounts for temporal constraints on motion, such as acceleration and velocity limits \cite{teb}. The constraints are weighted and optimized using a numerical optimization framework.

The poses of the robot and time deltas between poses are optimized in TEB.
Numerous constraints are considered in the problem formulation:

\begin{itemize}
  \item \textit{Waypoints} are attractive forces that encourage the robot to remain close to intermediate points along the original global path.
  \item \textit{Obstacles} (both static and dynamic) represent repulsive forces that keep the path from colliding with objects in its environment.
  \item \textit{Kinematic constraints} can be used to ensure that the optimized trajectory is kinematically feasible.
  \item \textit{Velocity and acceleration} constraints ensure the final trajectory is dynamically feasible
  \item \textit{Fastest path} constraints ensure that the optimized trajectory is temporally efficient.
\end{itemize}

In ROS~2, the TEB Local Planner package contains the TEB implementation \cite{tebgh}. 
It uses the {\tt g2o} framework to carry out numerical optimization \cite{g2o}.
At each control iteration, the poses from some portion of the global path and estimated time deltas between them are added to the graph, along with the constraints described above.
The graph is then optimized and the control command is extracted from the optimized TEB.
To help constrain the problem, the planner uses the previous cycle's optimized trajectory as the estimated starting values for the optimization in the current cycle.
Figure \ref{fig:teb_path} shows an example of an optimized TEB.
The final trajectory accounts for the kinematic constraints of the differential-drive platform being used, while also adhering to the original global plan. 

\begin{figure}[t]
    \centering
    \includegraphics[width=0.48\textwidth]{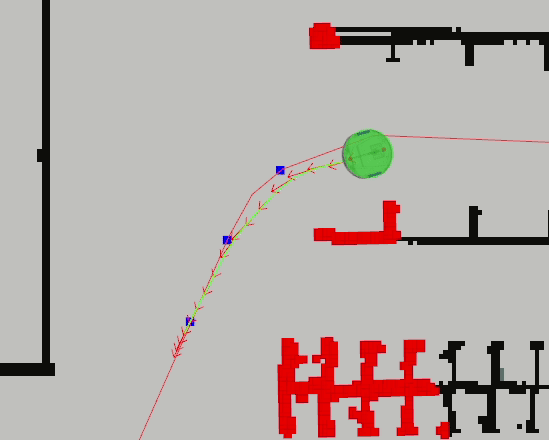}
    \caption{Example of an optimized Timed Elastic Band. The robot's global plan is shown as a red path. Way points are shown as blue squares. The robot's cost map is transformed into a vector representation (red blocks) that is suitable for distance calculations in the optimizer. The optimized trajectory, shown in green, comprises poses that are depicted as red arrows.}
    \label{fig:teb_path}
\end{figure}

The package is highly configurable.
As with other methods described, tuning these weighting parameters to produce optimal behavior for a given application can be time consuming.
Furthermore, as with all numerical soft-constraint optimization techniques, no guarantees can be given on the optimized solution, e.g., kinematic constraints may not be perfectly respected, or an insufficient weight on the obstacle distance constraint may result in the TEB skirting close to obstacles.
This is in contrast to sampling-based approaches like the Dynamic Window Approach, where kinematic constraints can be explicitly modeled during trajectory generation.

\textbf{Model Predictive Path Integral} is an MPC variant which uses perturbed trajectory samples to estimate the optimal trajectory \cite{mppi}.
Rather than numerically optimizing trajectories each time-step, a large number of noised samples are generated derived from the previous time-step's optimal trajectory, each is scored, and the optimal trajectory is identified.
Since this method does not involve the non-linear optimization stage found in most MPC techniques, the cost functions are uniquely not required to be differentiable or convex - yielding greater latitude in designing system behavior. 
The iterative update law optimizes the trajectory between time-steps, refining it as the robot moves towards its objective.
It has been proven effective in a number of domains including aggressive outdoor driving and commercial robotics.

In Nav2, the MPPI controller utilizes tensor representations to batch process trajectories.
The trajectory planner generates many randomly noised controls from Gaussian distributions for each control axis ($x$, $\theta$, and optionally $y$).
The noised controls are added to the previous optimal trajectory's velocities to create a set of randomly perturbed trajectories.
Constraints are applied on these trajectories' velocity limits and turning radii to make them physically achievable for omnidirectional, Ackermann, and differential-drive robots.
Using time samples in the trajectory, the velocities are then projected to fully construct the trajectory poses.

Each trajectory is scored by a configurable set of critic functions - or objective functions which optimize for particular elements of robot behavior.
Like DWB, all of the critic functions are dynamically loaded plugins and may be added, removed, or customized for a particular application.
The following critics are provided and produce good high-speed path tracking and active collision avoidance behavior: goal angle, goal distance, obstacle map, path align, path angle, path follow, prefer forward, anti-twirling, and dynamic constraints. 
After scoring, the batch's costs are normalized and a softmax function is applied to return the optimal trajectory. 
The first control in the optimal trajectory is then sent to the robot base controller to track.
This process is conducted iteratively to refine the optimal trajectory each time.
For the next time-step the trajectory is shifted, removing the first velocity which was sent to the robot base controller.
This improves performance by starting at the next predicted location with the remainder of the as-of-yet not executed portion of the trajectory. 
Thus, it is important that this controller runs at a sufficiently high rate with a sufficiently large number of samples. 
Empirically, 1000 samples at 50 Hz or 2000 samples at 30 Hz yield good results and are easily achievable in this CPU-only implementation (Table \ref{tab:controllers}).

MPPI behaviorally is a major improvement over reactive techniques.
It is exceptionally effective at reacting to dynamic obstacles intelligently even when not explicitly modeled as dynamic in the cost map.
Further, MPPI rarely requires active recovery behaviors to assist getting out of local minima due to becoming stuck or unable to make progress.
Its predictive back-out maneuvers when in close proximity to static or dynamic obstacles is a considerable asset in high-trafficked or confined environments to prevent the robot from becoming stuck.
Fig. \ref{fig:mppi} shows a similar behavior in isolation; whereas MPPI smoothly backs up and switches directions in a partial 3-point turn when a holonomic path begins in the opposite direction of the robot's initial heading.
While there are performance and behavioral differences between TEB and MPPI, the major relative strengths of MPPI is in its software architecture and test coverage for long-term maintenance, simpler and more intuitive configurations, and active maintainer presence. 

\begin{figure}[t]
    \centering
    \includegraphics[trim={1.4cm 2cm 1.4cm 5.5cm},clip,width=0.48\textwidth]{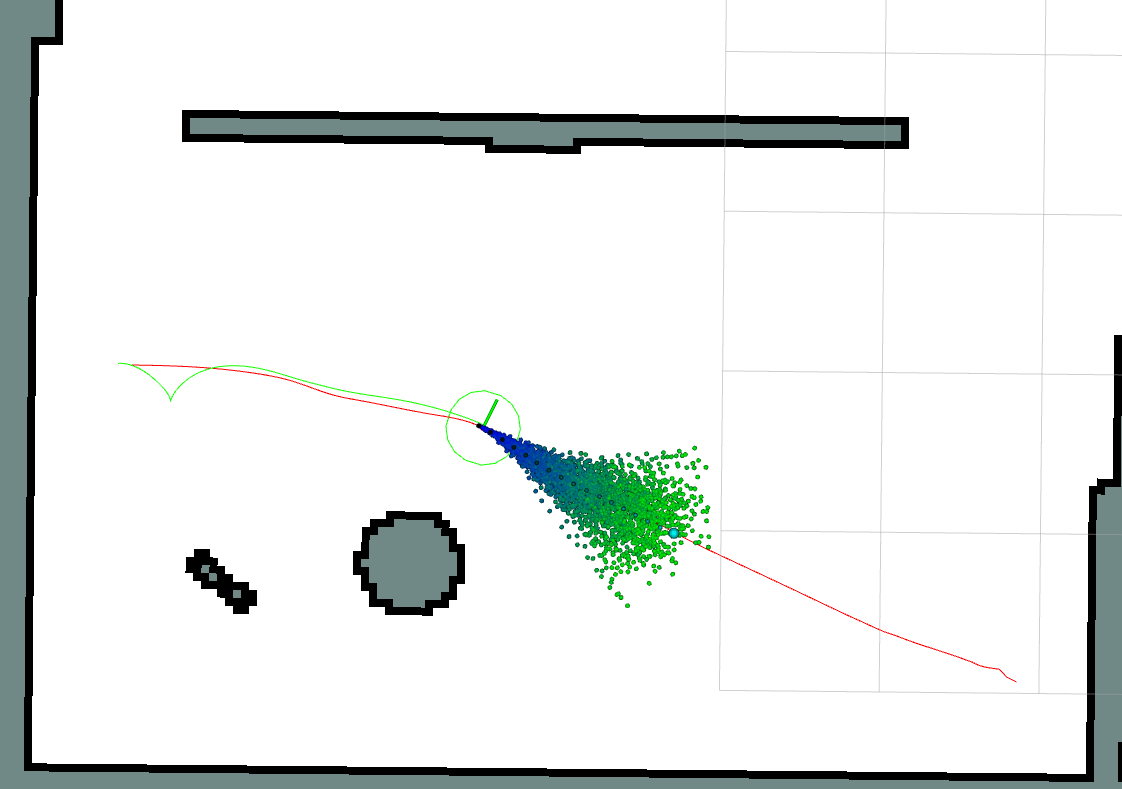}
    \caption{Example MPPI controller in action; red is the original global path, green is the actual path traveled by the robot (circle). The candidate trajectories points are shown, with the optimal selected trajectory highlighted on top.}
    \label{fig:mppi}
\end{figure}

\subsection{Geometric and Control-Law Controllers}

As opposed to reactive and predictive controllers, geometric and control-law derived controllers are quite simple. 
Their aim is to follow a given reference path or objective without deviation. 
Many of these techniques were early undertakings in trajectory planning, but have had an immense amount of staying power.
While dynamic behaviors give a robotic system some of its most directly visible intelligence, that is not necessarily optimal in some practical applications.
For example, exact route following behaviors have utility when following pre-planned routes through approved corridors in industrial settings where safety is paramount. 
Thus, it is an important type of behavior to support in a complete navigation framework.

These controllers do attempt to prevent collisions utilizing the cost map information given.
However, rather than adaptively avoiding collisions, these controllers will simply stop without going around an obstacle in its way.  
They will wait for a new global route to be selected or the obstacle moves from its path.
Due to their simplicity in evaluating geometric, kinematic, and/or control-law derived functions, these controllers can run well in-excess of 1 kHz. 

\textbf{Regulated Pure Pursuit} is a variation on the traditional Pure Pursuit algorithm. 
Pure Pursuit finds a point on the reference path a fixed-distance away from the robot to track, known as the \textit{lookahead point} \cite{Coulter1992ImplementationOT}.
This fixed-distance from the robot can be interpreted as the chord of a circle with points on it $(x,y)_{vehicle}$ and $(x,y)_{lookahead}$ belonging to the positions of interest.
When the path is represented in the vehicle coordinate frame, the curvature of the route, $\kappa$, to follow the lookahead point can be represented as $\kappa = \frac{2y}{L^2}$, 
where $y$ is the lateral coordinate of the lookahead point and $L$ is the distance away from the robot.
With an arbitrary choice of translational velocity $v_t$, we can compute an angular velocity, $v_a$, required to drive the robot towards the lookahead point via $v_a = v_t \kappa$.
This process is conducted iteratively resulting in the robot tracking a given reference path.


Adaptive Pure Pursuit builds on this by dynamically adjusting the lookahead distance based on the robot's current velocity, typically in the form $L = v_t l_t$, where $l_t$ is the lookahead time gain \cite{Kong2017AdaptivePP}.
This process has been shown to improve tracking stability of reference paths and reduce tracking errors in high curvature turns. 

The Regulated Pure Pursuit controller builds on these foundations and expands the behavior one step further with heuristic penalty functions on the desired translational velocity $v_t$ to \textit{regulate} the speed of the robot in confined or precarious situations \cite{rpp}.
Rather than naively assuming a fixed speed, which can be dangerous in practical applications, the robot is slowed when close to obstacles or the curvature of the turn exceeds configurable limits.
This slows the robot before sharp blind turns into partially observable settings (e.g. into or out of hallways and aisles) and reduces speed when close to other objects where the risk of collision is particularly high. 

This has been shown in experiments to decrease stopping distances for emergency stops, reduce overshoot and undershoot in high-curvature turns, all without substantially increasing time-to-goal metrics. 
The minimum velocity of the two heuristics is used, Eq. 3:

\begin{equation}
    v_{curv} = \frac{v_{t}}{r_{min} \; \kappa} \\
    \hspace{15pt}
    v_{prox} = 
 \begin{cases} 
      v_{t} \; \frac{\alpha \; d_{O}}{d_{prox}} & d_{O} \leq d_{prox} \\
      v_{t} & d_{O} > d_{prox} \\
   \end{cases}
   \hspace{28pt}
  \label{eq:aa}
\end{equation}

\vspace{10pt}

where $r_{min}$ is the minimum turning radius to apply the heuristic curvature heuristic, $d_O$ is the distance to the nearest obstacle, $d_{prox}$ is the proximity to obstacles to apply the curvature heuristic and $\alpha$ is a tunable weight.
Additionally, direction changes (e.g. forward to reverse) computed by feasible planners are used in RPP to switch directions as necessary at pre-planned poses.

\textbf{Graceful Controller} aims to provide smooth path following and automatic speed regulation for differential drive bases \cite{graceful}. This trajectory planner is best used with a kinematically feasible global planner.

This controller is based on a geometric control law derived for differential drive robot bases which generates only feasible commands \cite{park2011smooth}. The control implements a fast and slow subsystem. The slow subsystem is implemented as an Archimedean spiral that will take the robot from the current pose to a target lookahead pose, based on a single parameter, $k1$. The fast control subsystem then determines the linear and angular velocity required to follow the spiral based on the $k2$ parameter. The default values for $k1$ and $k2$ suggested by original paper appear to work well on a variety of platforms. The translational speed is automatically reduced from the maximum allowable speed when the curvature of the path tightens based on two parameters, $lambda$ and $beta$. While \cite{park2011smooth} contains the complete equations involved, higher values of $beta$ cause the overall drop in velocity to increase, while higher values of $lambda$ cause the rate of velocity reduction to be increased.

As with the RPP controller, the control law works on a single target pose (e.g. lookahead point). The controller works backwards through the global plan looking for the farthest point that is less than a maximum distance away from the robot and for which forward simulation of the control law to convergence results in no collisions. Selecting a pose too close to the robot can cause instability in the control law, so a minimum distance may also be defined. When no target pose can be found, the global planner is called upon to deliver a new path.

Unlike the RPP controller, the Graceful controller can produce smooth trajectories to non-trivial target poses without intermediate paths to follow due to its control-law based on spirals, Fig. \ref{fig:graceful}.
RPP and all Pure Pursuit variants are simple geometric approaches that requires continual updating of goal poses to produce path following behavior; for any single goal pose, computing a single $v$ and $\omega$ to drive towards it.

Thus, the Graceful controller is a relatively more intelligent path tracker that can accomplish a variety of other tasks (e.g. docking or object following) - however RPP still has great utility in exact path following behaviors, including respecting direction changes in feasible plans. 

\begin{figure}[ht]
    \centering
    \includegraphics[width=0.42\textwidth]{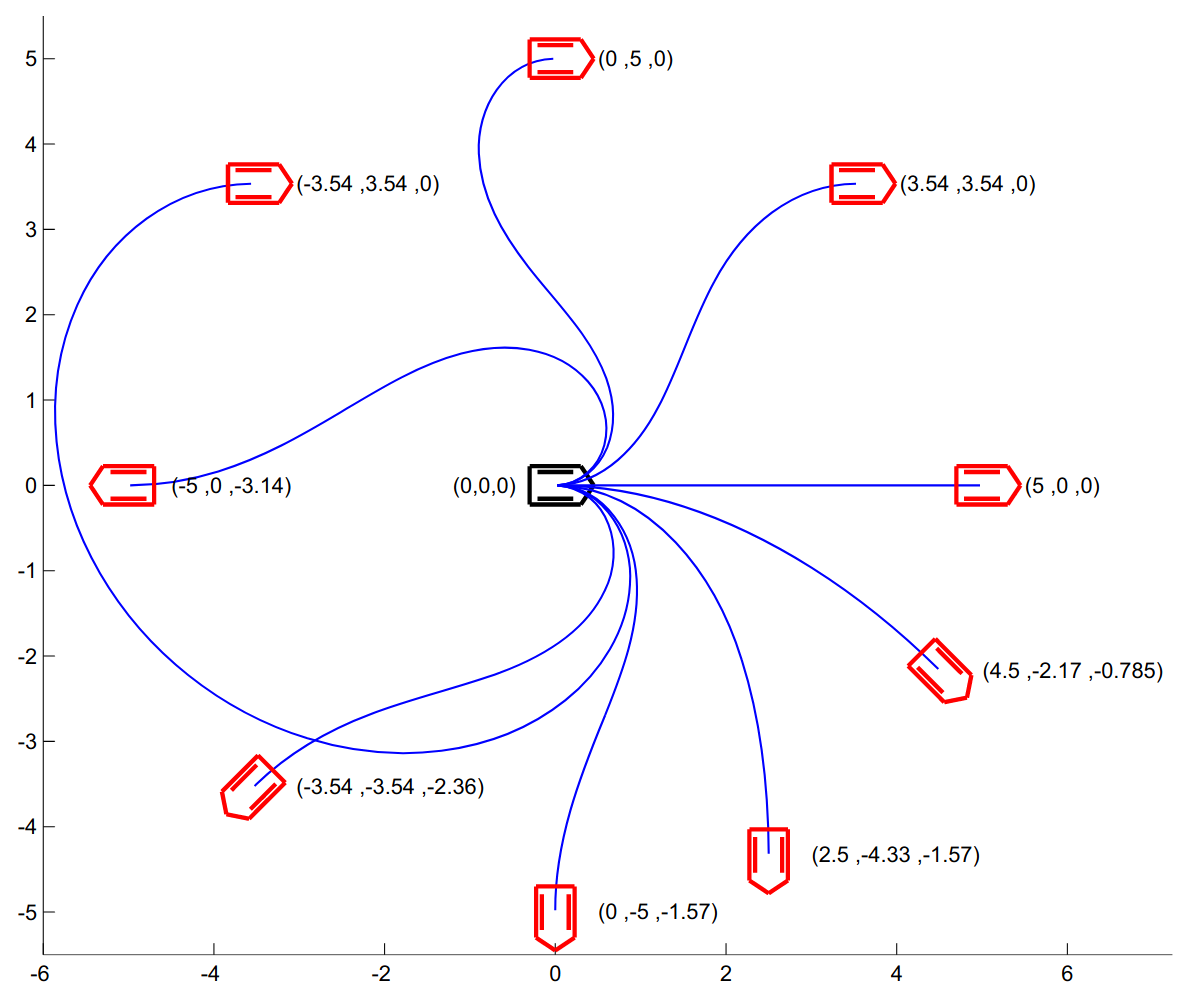}
    \caption{Graceful Controller's behavior achieving non-trivial local poses \cite{park2011smooth}.}
    \label{fig:graceful}
\end{figure}

A notable mode of operation is preferring final rotation. When enabled, the orientation of the final pose in the global plan will be replaced by an orientation that is tangent to the path. The robot will follow the path to the final pose and make a final in-place rotation to the desired heading. This behavior avoids a possibly large divergence from the planned path as well as large arcing behaviors.

Other features implemented in the controller include an orientation filter to clean up global plans and optional inflation of the robot footprint at higher speeds to increase safety margins (having a similar effect to RPP slowing when close to obstacles).

\textbf{Rotation Shim} is a re-usable component for robot platforms that may rotate in-place to rotate to a path's relative heading before beginning to track. 
This handles a frequent use case where controllers need to rotate a robot towards a newly planned path starting at another relative heading (e.g. from a holonomic planner from Sec.~\ref{sec:planners_infeasible}).
Once the robot's heading is within a configurable angular distance from the path's initial heading, the primary controller algorithm is used to track the path.
Controllers like DWB and MPPI are known to spiral out in a large arc towards the new heading or perform partial 3-point turns, respectively, when requested to track a path at a substantially different relative heading to its starting pose. 
This enables high-quality in-place rotations and removes the need for trajectory planners to process this special, but common and sometimes difficult initial task.

\section{Overview of Path Smoothing}
\label{sec:smooth}

\begin{table*}[!ht]
 \caption{Path Smoothers Comparison}
 \label{tab:smoothers}
 \begin{center}
 \begin{tabular}{ |c|c|c|c|c|c|c| }
   \hline
    & Time (ms) & Length (cm) & Ave. Cost & Max Cost & Smoothness & Ave. Turning Radius (m) \\
   \hline
   \hline
   Unsmoothed Path & -- & 1150.47 & 12.26 & 129.46 & 76.45 & 0.71 \\
   \hline
   Simple Smoother & 0.90 & 1115.39 & 18.89 & 132.89 & 58.79 & 3.07 \\
   \hline
   Constrained Smoother & 20.82 & 1149.37 & 6.88 & 103.98 & 87.85 & 2.59 \\
   \hline
   Savitzky–Golay Smoother & 0.06 & 1146.75 & 12.18 & 127.76 & 76.19 & 2.29 \\
   \hline
 \end{tabular}
 \end{center}
 \hspace{60pt} An inter-modal algorithm selection summary can be found in Appendix \ref{algo_selection}.
\end{table*}

New in Nav2 are smoothing capabilities to refine global paths.
Smoothers have long been employed in sampling-based planning algorithms to interpolate between points and smooth out irregular movements. 
While plans produced by search-based planners are typically less tortuous, many can be improved with additional refinements: 2D-A* can smooth jagged motions due to Moore search neighborhoods, Theta* can obtain continuous transitions where line segments meet, and feasible planners can benefit from curvature minimization. 

Canonically, local trajectory planners have been successful in producing relatively smooth behavior from unprocessed global paths while considering dynamics and other constraints.
However, improving path quality can further reduce jitter in geometric and control law-based path tracking, as well as improve stability for predictive and reactive algorithms \cite{smooth_paths}. 
Thus, Nav2 provides three path smoothing algorithms to support holonomic and feasible planners - the Simple Smoother, Constrained Smoother, and Savitzky-Golay Smoother.

Table \ref{tab:smoothers} show a comparison of the path smoothers, where the costs reported are  those in the cost map and smoothness is measured as $|\Delta x_{i+1} - \Delta x_{i}|$, ($\Delta x_i = x_i - x_{i-1}$).
These are the results of an experiment composed of 1,000 randomly generated start and goal poses in a representative cluttered home environment (e.g. furniture, hallways, open rooms, random objects strewn about) using the Smac Hybrid-A* planner.
These paths were then smoothed using the algorithms in this section and compared on a set of key metrics: compute time, path length, cost or distance from the nearest obstacle, smoothness, and path curvature.

\textbf{Simple Smoother} is used to remove localized imperfections such as oscillation, discontinuities, and turning transitions in the holonomic planners, typically in 1-6 ms.
As its purpose is to remove localized irregularities or discontinuities, it uses a simple gradient descent formulation which minimizes the objective function:

\begin{align}
    \sum_{i=1}^{N - 1} \omega_s (\Delta x_{i+1} - \Delta x_{i})^2 + \omega_d (x_i - y_i)^2
    \label{simple}
\end{align}

where $\omega$ are configurable weights, $x$ is the smoothed path and $y$ is the original path.
The first term is commonly used and optimizes for smoothness, while the second term penalizes adjusting the smoothed path too far from the initial path.
Together, they provide smoothed behavior in the local neighborhood of the unsmoothed path resolving defects such as sharp turns, jagged regions, and odd characteristics produced by gradient descent on discrete grid maps.
This is born out in the experimental results - this smoother is quite fast and generates improved paths.
However, because this will smooth out turning corners, it may increase the path costs relative to the initial plan as the smoothed result becomes closer to obstacles (ex. Fig \ref{fig:SimpleSmoother}). 

The newly optimized path has its via-point orientations assigned as the tangent to the path.
The smoothing terminates when the parameter tolerance is achieved or the maximum time has been used.
If a path is found to be in collision, the last collision-free path from prior iterations is returned.
Fig. \ref{fig:SimpleSmoother} illustrates the Simple Smoother for raw paths produced by Theta* and Smac 2D-A*.

\begin{figure}[ht]
    \centering
    \includegraphics[width=0.48\textwidth]{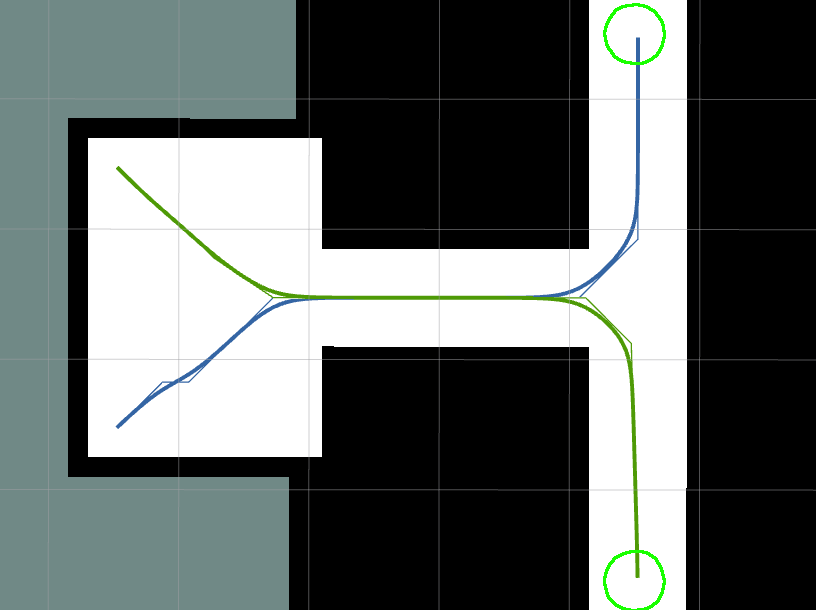}
    \caption{Theta* and Smac 2D-A* paths smoothed by Simple Smoother. Theta* depicted in green, Smac 2D-A* - in blue. Non-smoothed paths depicted in narrow lines, smoothed - in bold.}
    \label{fig:SimpleSmoother}
\end{figure}

\textbf{Constrained Smoother} is designed for refining the feasible planners from \ref{sec:planners_feasible} with a more developed optimization cost function taking into account curvature, cost fields, and smoothness.
It derives from the smoothing algorithm introduced in Hybrid-A*, but rather than using a Voronoi Field, it leverages the existing cost grid with obstacle inflation \cite{dolgov2008practical}:

\begin{align}
    \begin{split}\label{constrained}
    {}& \omega_{s}  \sum_{i=1}^{N - 1} (\Delta x_{i+1} - \Delta x_{i})^2 + \omega_{c} \sum_{i=1}^{N - 1} (cost(x_i, y_i))^2\\
    & + \omega_{\kappa}  \sum_{i=1}^{N - 1} \sigma_\kappa (\frac{\Delta\Phi_i}{|\Delta x_i|} - \kappa_{max})\\
    \end{split}
\end{align}

\begin{figure}[ht]
    \centering
    \includegraphics[width=0.48\textwidth]{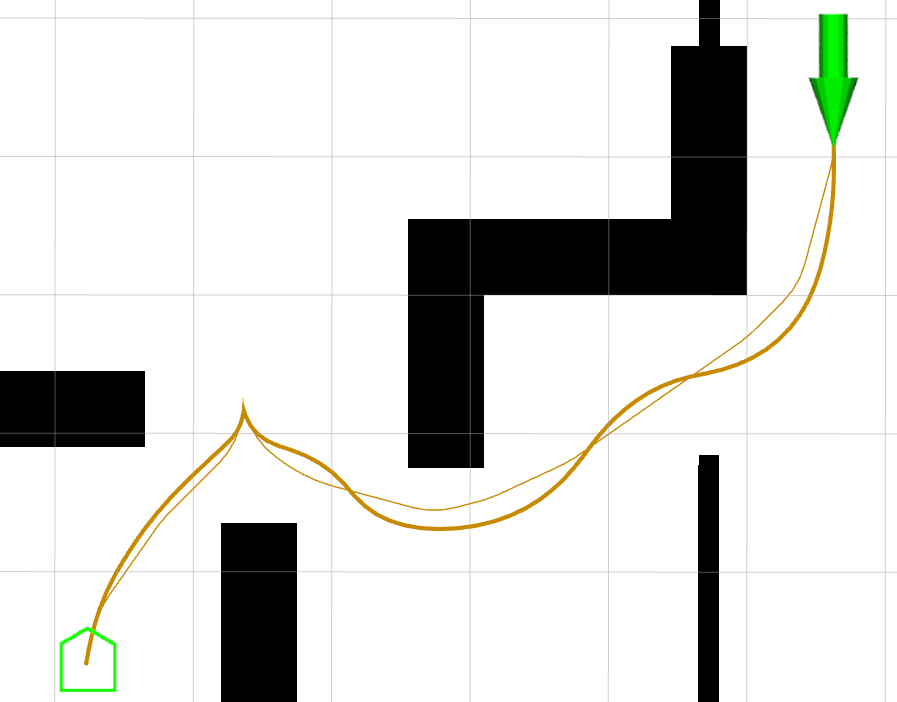}
    \caption{Smac Hybrid A* path smoothed by Constrained Smoother. Non-smoothed path depicted in narrow line, smoothed - in bold.}
    \label{fig:ConstrainedSmoother}
\end{figure}

where $\omega_{s}$, $\omega_{c}$, and $\omega_{\kappa}$ are the weights for the smoothness, cost, and curvature terms, respectively; $\sigma_{\kappa}$ is a quadratic penalty function, $\kappa_{max}$ is the maximum feasible curvature, $\Delta\Phi_i$ is the change in tangential angle, and $cost(x_i, y_i)$ is the cost at a given cell.
This objective function is solved using Google Ceres, providing a number of linear solver options.

This provides more comfortable paths for feasible planners by minimizing curvatures (from their maximums in search) to improve practical drivability while retaining kinematic feasibility.
It also balances improving the smoothness of the path with shifting the plan into lower-cost regions. 
Since it has no constraint on the original path coordinates, this smoother may make larger, global level changes to the path beyond localized artifacts.
Thus, the benchmark shows that the Constrained Smoother generated paths typically longer then the Simple Smoother, but with significantly reduced costs, while also improving on smoothness and curvature from the original path.
Fig. \ref{fig:ConstrainedSmoother} demonstrates this visually with a path of similar length, but maximizing distance from obstacles and reducing path curvature.

When reversing path segments exist, $\omega_{cc}$ (cusp-cost) is used instead of $\omega_c$ within close proximity to cusp points.
This value is strictly larger than $\omega_c$ and incentivizes the optimization problem to push direction-changing maneuvers further away from obstacles for additional safety margins. 
Figure \ref{fig:wcc} shows an example of the impact of an elevated $\omega_{cc}$ in close proximity to cusp points.

\begin{figure}[ht]
    \centering
    \includegraphics[angle=90,width=0.48\textwidth]{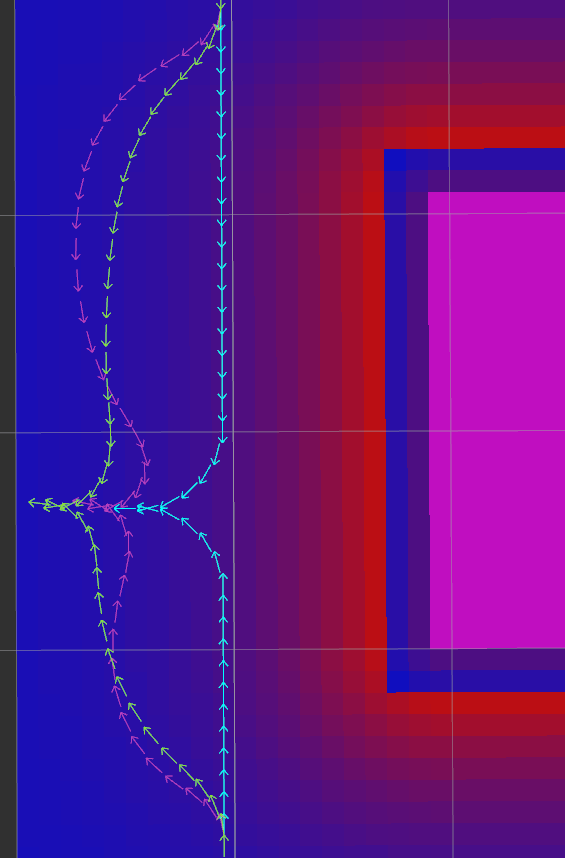}
    \caption{Impact of increased cusp costs near direction-changing points. The cyan path is the unsmoothed original path, purple is using the same $\omega_{c}$, and green uses $\omega_{cc}$ near the direction change. The cost-field is shown as the background colored gradient near a rectangular obstacle \cite{wcc_tix}.}
    \label{fig:wcc}
\end{figure}

While the Constrained Smoother is targeting feasible planners, it may also be used by the holonomic planners.
The curvature term is removed when $\omega_\kappa = 0$ and may operate on paths using only the cost and smoothness terms to perform global-level optimization rather than only removing localized imperfections and discontinuities.
However, this smoother is much more computationally expensive than the Simple Smoother, taking up to several seconds depending on the length of the global path and the environment.
Thus, it is primarily recommended where maintaining feasibility is a strict requirement. 
The smoother terminates when the parameter, function, or gradient tolerance is achieved, the maximum number of iterations has been reached, or the maximum time has been used.

\textbf{Savitzky-Golay Smoother} is a path smoother derived from the famed digital filter of its namesake \cite{savitzky64}.
This filter analytically smooths regularly-spaced data using a sliding window polynomial function to remove noise from the signal. 
This may be applied to path plans from search-based planners due to their regular search intervals. 
This smoother is unique from the others in that it will not generally modify the character of the input path. Rather, it will only smooth out irregularities that can be modeled as noise within a 7-point window.
This is most useful for gradient-descent path creation, as found in the NavFn Planner or potential field methods, which can generate irregularities due to performing gradient-descent on a discrete grid.  
This method will smooth "noisy" outlier path points while retaining the intent from the planner.
Fig. \ref{fig:SavitzkyGolaySmoother} shows a case in which NavFn generated a path with an unusually large numbers of jagged points, which the Savitzky-Golay Smoother refines into a smooth global path.
Since the character of the path remains the same, the changes in key metrics are often negligible.
This smoother, like the others, will also update the orientation vectors of each path point to be tangent to the final smoothed path. 

\begin{figure}[ht]
    \centering
    \includegraphics[trim={7cm 0 5cm 1cm},clip,width=0.48\textwidth]{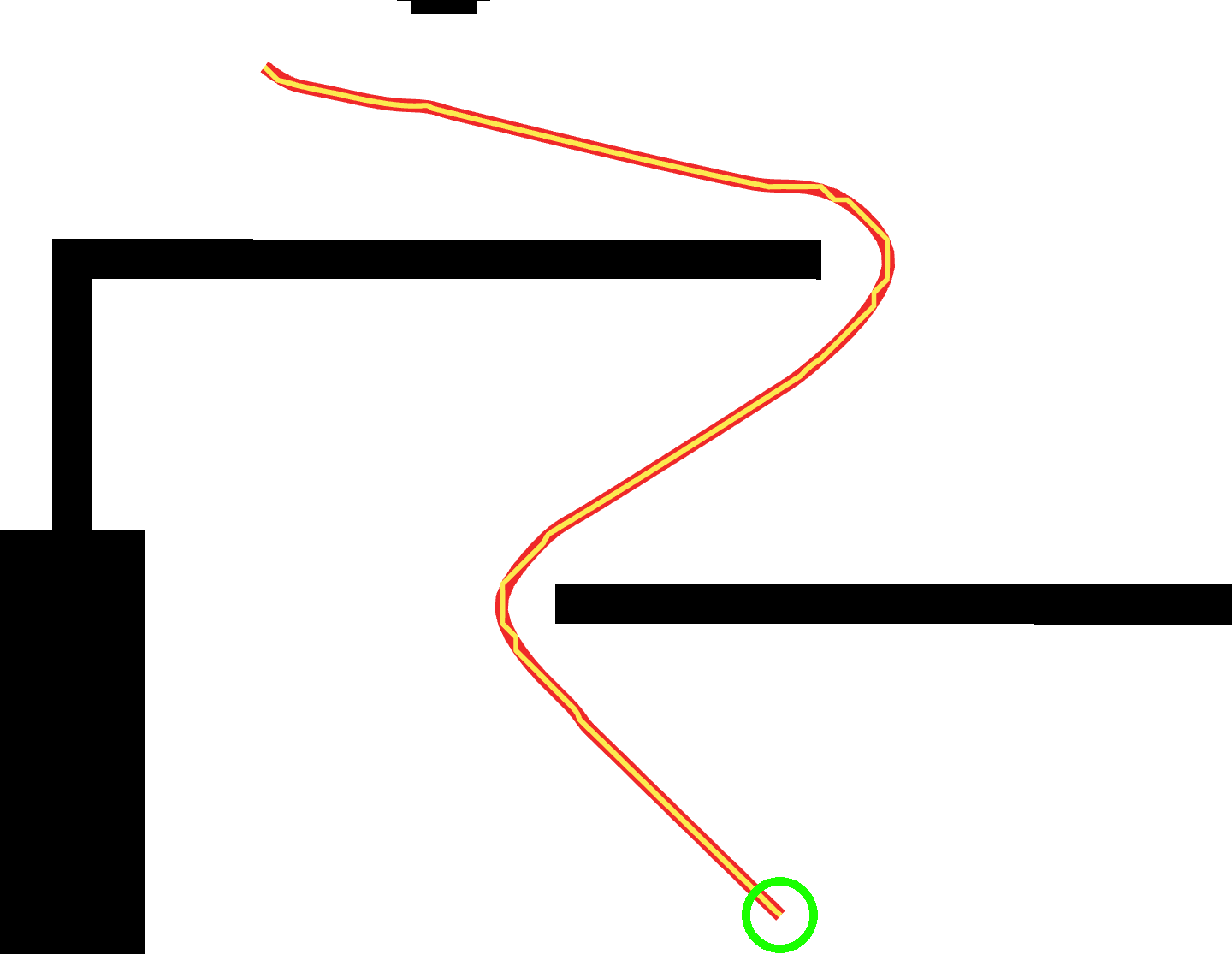}
    \caption{NavFn path (yellow) smoothed by Savitzky-Golay Smoother (red).}
    \label{fig:SavitzkyGolaySmoother}
\end{figure}

\section{Overview of Perception}
\label{sec:costmap}
\subsection{Overview}
The key to achieving reactive behavior - where a robot intelligently navigates in a dynamic environment - is using the robot's sensors and semantic information to maintain a representation of the environment.
ROS~2 employs \textit{cost} or \textit{risk maps} as the environmental model to consolidate the results of potentially many algorithms.
A balance must be achieved to create a sufficiently high-fidelity representation of the environment and be able to maintain that representation efficiently.
Perception algorithms, sensor data, or semantic information about the environment are populated into this world model for use in global path and local trajectory planning.

Cost maps reduce the world model into a two-dimensional grid, with each grid cell having a cost associated with it.
Cost maps evolved from the probability-based occupancy grids developed at Carnegie Mellon in the 1980s \cite{matthies1988,moravec:grids}.
The modern cost map was first described by Marder-Eppstein et al \cite{nav}, where the cost associated with each cell was an integer in the range [0, 255]. 
In general, planning algorithms will prefer paths with lower costs and avoid those with higher costs, but there are also special values for unknown and occupied cells (255 and 254, respectively) to provide hard constraints on robot behavior.
These special values allow planning and collision checking algorithms to guarantee viable navigation in only free and known regions. 
The cost map acts as the planning algorithms' configuration space \cite{lozano1983spatial}.
Figure \ref{fig:Costmap} shows an example of the key values in a cost map.

The cost map provides a reasonably balanced approach to the fidelity/efficiency trade-off on both low and high power compute platforms. 
The grid data structure provides a simple correspondence with the real world and also provides constant time lookup. 
One drawback is that the memory consumption required to maintain the cost map scales linearly with the number of cells. 
The use of Quad-trees to represent occupancy grids has been proposed to alleviate this scaling via multi-resolution data storage \cite{quadtree}.
However, these methods fail to recover substantial memory benefits when cost map convolution or inflation are applied, or when the density of obstacles in the environment leaves little open space to represent with coarser-resolution quantizations - while also introducing non-constant look-up times.
While these are useful developments for some applications, it is most common for users in ROS~2 to apply a form of cost map inflation and operate in restricted spaces. 
In practice, the size of the cells in the cost map is 0.05 meters by 0.05 meters, which typically is small enough to render the environment with sufficient fidelity, but also large enough to not require enormous amounts of computation. 

While a two-dimensional cost grid discards a great deal of potentially useful information, it has been applied repeatedly to solve a great number of practical robotics challenges.
Further, it is common for perception algorithms populating the cost map to maintain higher fidelity representations internally for modeling the environment, which are binned or reduced to two dimensions for consolidation with other algorithms.
Maintaining a two-dimensional cost map enables the application of a broad number of highly-efficient planning, control, and collision checking techniques for arbitrary types of robot systems. 

The process by which the cost map is updated and maintained in ROS~2 is the \textit{Layered Costmap} \cite{costmap}: an extensible, systematic way to update the cost map with many, configurable, and arbitrary perception algorithm results.
The Layered Costmap comprises an ordered list of dynamically run-time loaded layers which each represent a different data source, algorithm, or result.
Each update cycle, the main cost map is updated by polling each of these layers in turn to populate the grid with new information.

Any developer may write a costmap layer and include it on their robot system for their specific applications.
Common types of layers include: cost map inflation, sensor processing, semantic information, results from machine learning detection or segmentation algorithms, and multi-robot coordination. 
However, a set of common layers are provided by ROS~2 applicable for a broad range of common applications. 

\begin{figure}[ht]
    \centering
    \includegraphics[width=0.48\textwidth]{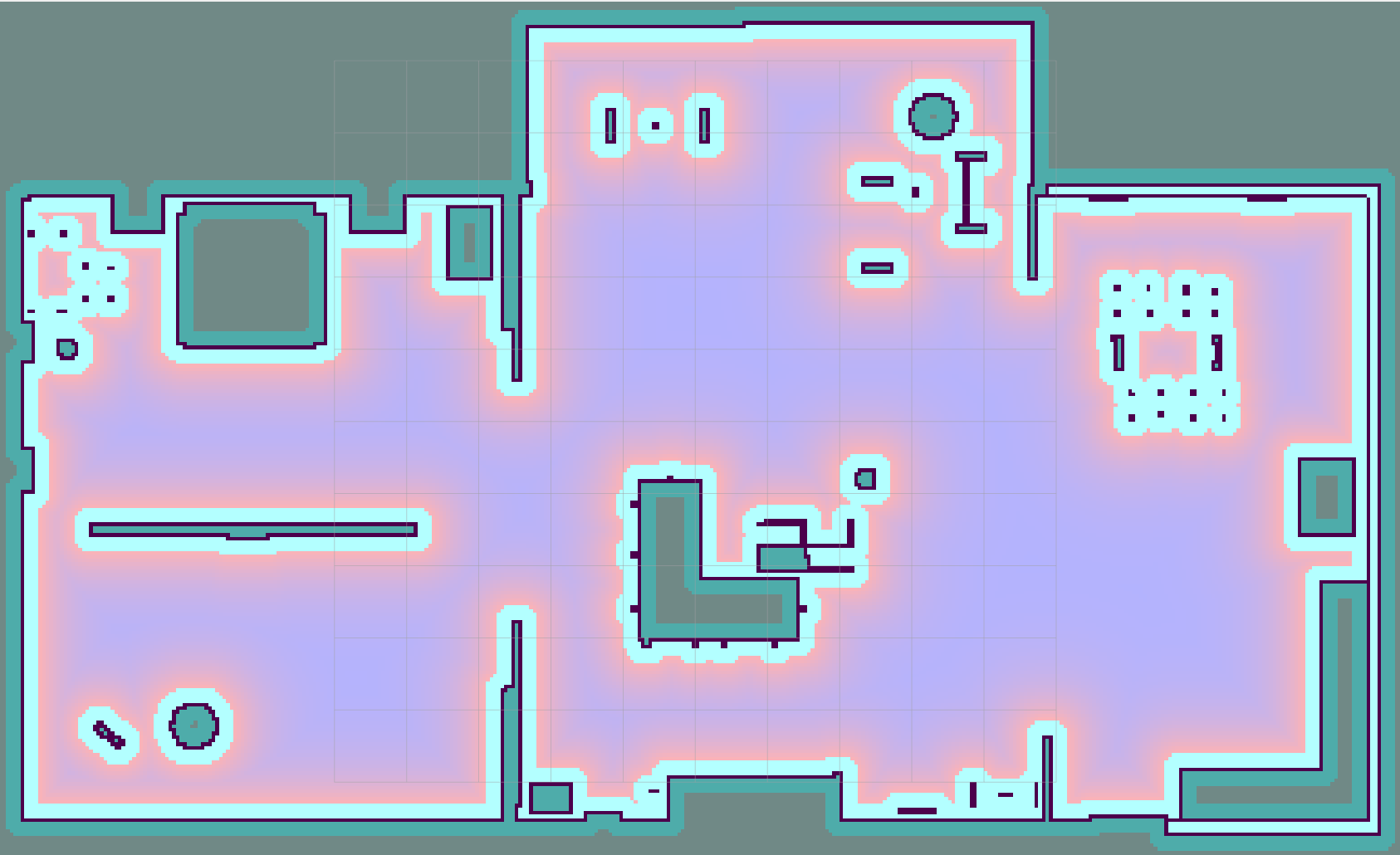}
    \caption{Cost Map example in a practical home environment. Black are lethal obstacles, light blue are inscribed lethal costs due to robot dimensions, the gradient from red to blue are a cost potential field caused by the Inflation Layer, and grey are unknown cost cells. Free space with no potential cost are white, not shown.}
    \label{fig:Costmap}
\end{figure}

\subsection{Costmap Layers}
\textbf{Static}
A map provides information about where obstacles are known to be a priori, often from an offline mapping algorithm or other data source like blueprints. 
The static layer subscribes to an \texttt{OccupancyGrid} topic containing this information.
It is common for the static layer to be the first costmap layer, providing the base information about the environment for later algorithms to modify.
While the name implies non-changing, the static layer can have the map change over time as the result of map sharding over large spaces or updates to the environment due to continual mapping.
There is a separate subscription to a \texttt{OccupancyGridUpdate} topic for efficient non-dimension changing updates to localized map information. 

\textbf{Obstacle}
The obstacle layer stores information from high accuracy sensors such as lasers and RGB-D cameras in a two dimensional grid. Each point in the sensor data is treated as a ray, originating from the sensor pose.
That ray is ray-traced through the grid in two dimensions using Bresenham's algorithm, where the endpoint is marked as occupied and its path is marked as free space due to direct visibility.
Fig. \ref{fig:raytracing} demonstrates this process visually.
The 2D laser scanner produces measurements from obstacles which are marked in the grid in blue. 
Breshenham's ray-casting algorithm is used to clear out the visible free-space from the sensor to the obstacle in white, while the remainder of the grid is still unknown (grey). 
Due to the high accuracy of the sensors, these observations are treated as absolute, without the probabilistic calculations of occupancy seen in the 1980s. 
The simplification of logic also assists in quicker computation.
As a robot moves through the scene and/or more measurements are taken, it will dynamically generate an occupancy map of the environment around it based on depth sensing.

The two dimensional nature of this layer makes it ideal for planar laser scanners, but may be overly constrained when using three-dimensional data.
It may be necessary when working with low-accuracy or noisy sensors to pre-filter the data stream before processing in the obstacle layer to reduce noisy results in the cost map. 

\begin{figure}[ht]
    \centering
    \includegraphics[width=0.48\textwidth]{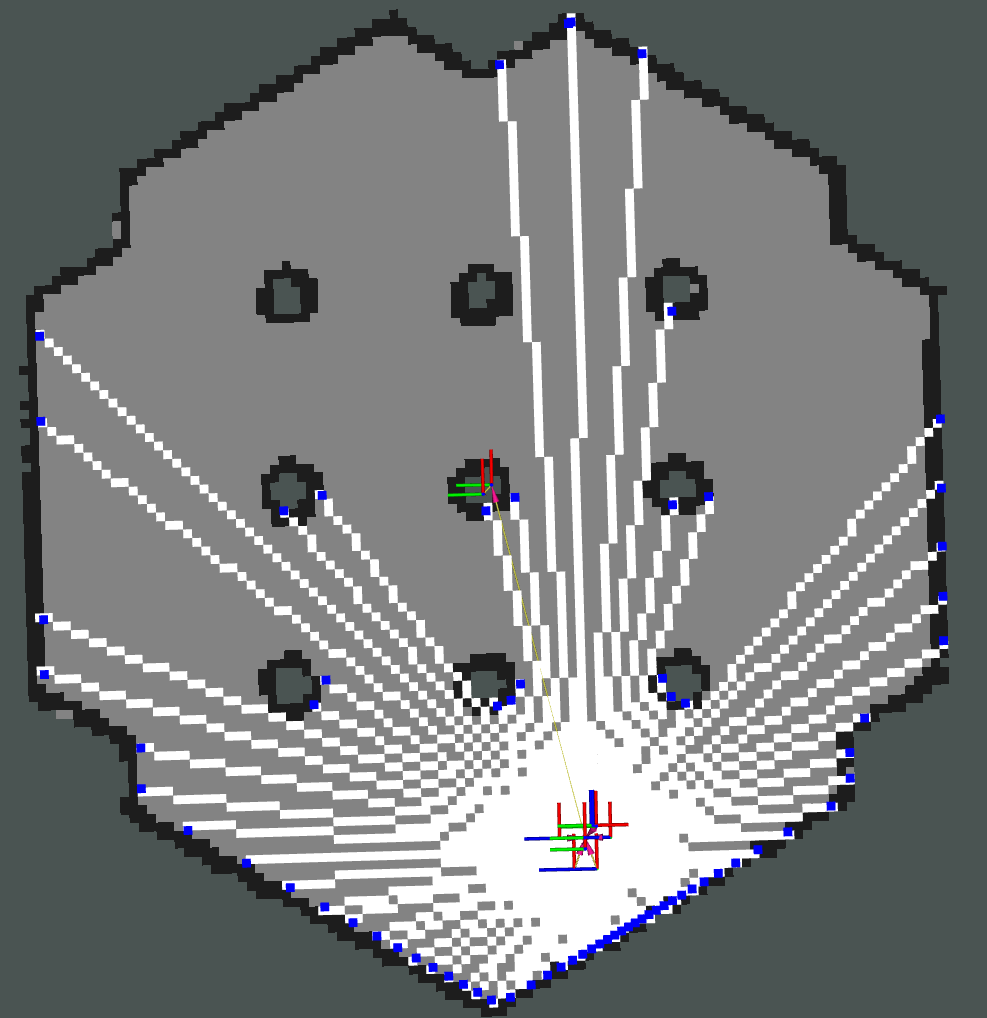}
    \caption{Visual demonstration of ray-tracing based marking and clearing in a grid map. Overlaid on a sandbox map, a low-resolution laser scan is raytracing freespace (white) from measurements (blue).}
    \label{fig:raytracing}
\end{figure}

\textbf{Voxel}
The voxel layer is similar to the obstacle layer, but tracks the environment and ray-traces measurements in three dimensions. It is most useful for RGB-D, 3D laser scanners, or other non-trivially two-dimensional sensor streams.
Internally, it uses a voxel grid model the size of the cost map to represent the environment in a non-probabilistic way.
This model scales linearly with cost map size, similar to the two-dimensional grid, by maintaining the height dimension through manipulating the bits of an unsigned 32 bit integer. 
Thus, its maximum height is limited.
It is recommended to only track heights to the limit of what is useful (e.g., a home vacuum robot need not maintain a voxel grid $> 1 m$ off the ground).
The vertical resolution of the voxels is independent from their horizontal resolution, and is typically larger. 
The three-dimensional data is projected down to two dimensions for writing into the main cost map.
As it conducts three-dimensional raytracing, this layer is more computationally expensive than the obstacle layer.

\textbf{Non-Persistent Voxel}
The non-persistent voxel layer computes values in the same manner as the voxel layer, but does \textit{not} persist the data from one update to the next. 
Thus, each iteration a new voxel grid is created to populate the cost map, then cleared.

\textbf{Spatio-temporal Voxel}
The Spatio-temporal voxel layer (STVL) use a more refined data structure than the voxel layer, Fig. \ref{fig:stvl}.
It tracks obstacles in three-dimensions using OpenVDB, a sparse voxel grid library developed by Dreamworks Animation for movies such as \textit{How to Train Your Dragon} and \textit{Puss in Boots} \cite{openvdb}.
Rather than ray-tracing to clear obstacles in the voxel grid, STVL uses a method called \textit{decay acceleration} to maintain a representation of the environment based on the sensor's measurement frustum and current measurements.
As measurements are processed, the timestamp of the data is added to the voxel belonging to the reading.
If an active (e.g. occupied) voxel is within the sensor frustum and seen by new data, the time is updated.
If a voxel is not viewed by new data, a decay acceleration factor is applied to its time, quickening its removal due to lack of current visibility.
The frustum models are configurable and currently support 3D lidar 'donut' models and RGB-D, depth, and similar sensors using a 6-plane bounding volume defined by the minimum and maximum range and horizontal and vertical field of views.

For all voxels outside of the sensor's frustum, a global linear time decay is applied, usually 10-30 seconds.
After voxels are expired due to either method, they are removed from the data structure.
Thus, its storage size is only dependent on the number of currently active and occupied cells.
The sparse voxel data structure is projected to two dimensions for inclusion in the main cost map.
This layer is most appropriate for robots operating in highly dynamic environments with a high degree of sensor coverage.
It was created to resolve issues with discrete cell ray-tracing which occasionally left uncleared obstacle cells in highly dynamic environments with many moving agents. 

\begin{figure}[ht]
    \centering
    \includegraphics[width=0.48\textwidth]{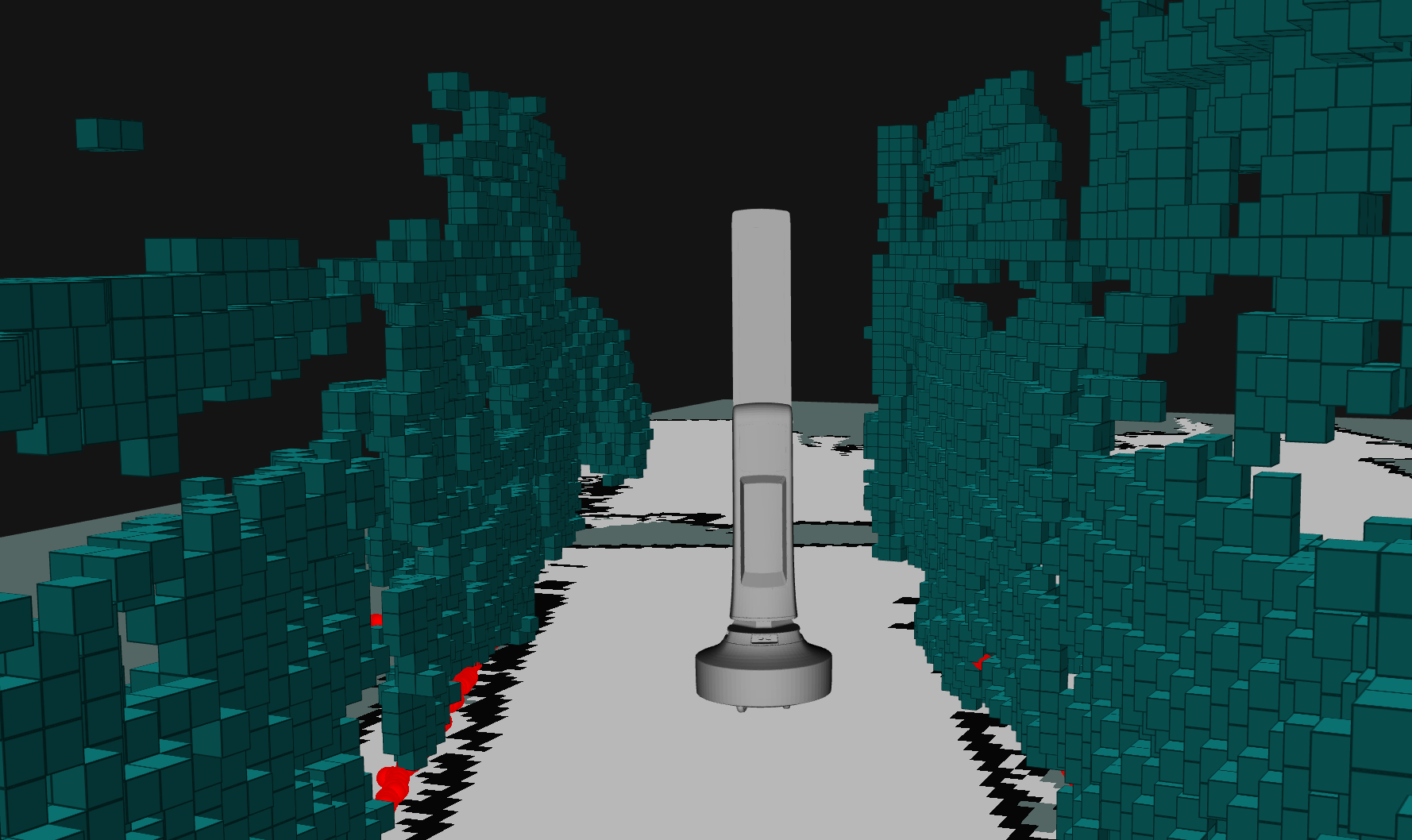}
    \caption{STVL's internal sparse 3D voxel grid representing shelves in a retail environment \cite{stvl}.}
    \label{fig:stvl}
\end{figure}

\textbf{Range}
The range layer processes data from Infrared, Sonar, and Ultrasonic sensors.
Due to the noise inherent to these types of sensors, the range layer tracks occupancy using a probabilistic model in which only cells with high confidence predictions are written into the main cost map. 
It has support for both fixed reading (e.g. binary detection of obstacles) or variable reading sensors.
The probability of an obstacle in a cell is calculated using the static properties of the sensor (range of view, maximum reliable range and sensor variance), the sensor reading itself, and the angle and distance of the cell relative to the sensor origin \cite{oriolo1997fuzzy}.


\textbf{Inflation}
The inflation layer uses the occupancy information from the other layers to add an inflated exponential decay function around untraversable regions.
It does this by iteratively expanding from known obstacles to a configurable distance, applying Eq. \ref{inflation}:

\begin{equation} \label{inflation}
  cost(x, y) = (cost_{lethal} - 1) \hspace{3pt} e^{-\omega_{scale} \hspace{2pt} (d_o - r)}
\end{equation}

where $\omega_{scale}$ is the cost scaling factor, $d_o$ is the distance to the obstacle, $cost_{lethal}$ is the lethal obstacle cost, and $r$ is the inscribed radius of the footprint.
For circular robots, the cost map regions that are traversable after this is applied represent the configuration space of the robot.
For non-circular robots, the cell cost for the largest distance between an edge of the robot and the center can be used with Eq. \ref{inflation} to recover the cost at the center where full footprint collision checking is required. 
When the center cost is below that value, it is known that the robot cannot be in collision. 
Thus when using non-circular robots, it is important to still use inflation to optimize collision checking used in such algorithms as found in the Smac Planner framework described in \ref{sec:planners_feasible}. 
It is recommended to set a smooth inflation decay across the cost map, rather than maintaining only steep costs near obstacles, as shown in Fig. \ref{fig:Costmap}.

\textbf{Keepout}
The keepout layer (a.k.a. keepout filter) uses input masks to mark cells as obstacles.
These masks are highly configurable to contain a great deal of semantic information and interpreted through {\tt CostmapFilterInfo} into cost values to apply to the grid.
These need not be only binary obstacle information, but also setting a range of costs to weight certain regions over others.

\textbf{Speed}
The speed limiting layer uses the same filters API as the keepout zones, but embedding maximum speeds as either percentages or absolute values in the filter masks. 
This layer allows users to specify zones at which a robot should slow down, such as when entering or exiting aisles.  

\textbf{Binary}
The binary layer uses the same filters API as previously introduced, but embeds binary behavioral triggers into a spatial mask.
When the cost value crosses the threshold, a ROS message is sent to a client server to indicate a change in binary state.
An illustrative example would be to set spatial bounds conveying locations where it is and is not appropriate to stream camera feeds to a robot operations center.
There may be regions of homes (such as bathrooms) and other facilities where remote monitoring of camera data would breach legal or ethical bounds.
This layer allows the native embedding of spatial constraints to trigger binary actions.

\section{Overview of Behavior Trees}
\label{sec:bt}

In this section, we describe the architecture of how robotic components previously described are assembled to create autonomous navigation.
Configurable behavior trees (BTs) are used to orchestrate internal robot navigation tasks such as planning, control, high-level behaviors, recoveries, and more in Nav2 \cite{nav2,bt}.
This contrasts historical uses of state machines and hard-coded logic for designing navigation systems: typically passing a path to a path tracker with hand programmed behaviors \cite{nav}.
While it is possible to model most BT configurations with hierarchical state machines (and vise versa), behavior trees have been the subject of much interest and are increasingly preferred from the results of user studies.
The \textit{BT Navigator} is the highest level component of Nav2, taking requests and processing them via the behavior outlined in the tree.
Each behavior tree is modeled as an XML file comprising many BT nodes which are dynamically loaded at run-time.
The behavior tree can be specified per-run so a system may take on different behaviors or tasks.

Most BT nodes will invoke a remote server over a ROS~2 interface (action, topic, service) to compute a value for component-level modularity, although this is not required.
A growing number of primitive BT nodes are provided by Nav2, shown in Table \ref{tab:behavior_trees}, that may be used to develop useful robot behaviors.
Note that users may easily create and use their own BT nodes and trees.

\begin{table}[!ht]
 \caption{Included Behavior Tree Nodes}
 \label{tab:behavior_trees}
 \begin{center}
 \begin{tabular}{ |c|c|c|c|c|c|c| }
   \hline
   Behavior Tree Node & Description  \\
   \hline
   \textbf{Action} &   \\
   \hline
   Navigate To Pose & Navigate from A to B  \\
   \hline
   Navigate Through Poses & Navigate from A-Z through B...Y  \\
   \hline
   Compute Path To Pose$^\diamond$ & Plan from A to B  \\
   \hline
   Compute Path Through Poses$^\diamond$ & Plan from A-Z through B...Y  \\
   \hline
   Follow Path$^*$ $^\diamond$ & Follow Path   \\
   \hline
   Clear Costmap & Clear environmental model   \\
   \hline
   Back Up$^*$ & Reverse  \\
   \hline
   Spin$^*$ & Rotate \\
   \hline
   Drive On Heading$^*$ & Drive forward \\
   \hline
   Smooth Path & Smooth path \\
   \hline
   Remove Passed Goals & Remove goals passed B...Y \\
   \hline
   Truncate Path & Truncate path near goal \\
   \hline
   Wait$^*$ & Pause \\
   \hline
   Reinitialize Localization & Reset localization  \\
   \hline
   Assisted Teleop$^*$ & Tele-op with collision avoidance  \\
   \hline
    \textbf{Condition} &  \\
   \hline
    Distance Traveled & If a distance has been traversed \\
   \hline
    Goal Reached & If the goal is reached \\
   \hline
    Goal Updated & If the goal has changed \\
   \hline
    Initial Pose Received & If initial pose is set \\
   \hline
    Is Battery Low & If battery too low to continue \\
   \hline
    Is Path Valid & If path is not in collision \\
   \hline
    Is Stuck & If not making progress \\
   \hline
    Is Transformation Available & If transform is possible \\
   \hline
    Is Time Expired & If time has passed \\
   \hline
    Path Expiring Timer & Timer before last path is stale \\
   \hline
    \textbf{Decorator} &  \\
   \hline
    Rate Controller & Control flow by set frequency \\
   \hline
    Distance Controller & Control flow if distance has passed \\
   \hline
    Speed Controller & Control flow proportional to speed \\
   \hline
    Single Trigger & Trigger control flow once \\
   \hline
    Path Longer On Approach & If path too long on goal approach \\
   \hline
    Goal Updated & Control flow if new goal set \\
   \hline
    Goal Updater & Bypass and update goal over topic \\
   \hline
    \textbf{Control} &  \\
   \hline
    Pipeline Sequence & Flow-of-water sequence \\
   \hline
    Recovery & Executes second node if first fails \\
   \hline
    Round Robin & Ticks children in round-robin style \\
   \hline
 \end{tabular}
 \end{center}
 \hspace{25pt} $^*$ Includes complementary cancel task node 

 \hspace{25pt} $^\diamond$ Includes complementary algorithm selector node
\end{table}

Nav2 provides a number of behavior trees comprising these nodes.
These fall into three major categories: navigate to pose, navigate through poses, and task-specific applications.
Each have several customizations available to provide intelligent behavior in various situations.

The navigate-to-pose behavior trees provide basic pose-to-pose navigation.
The BTs have support for consistent replanning at a fixed frequency\footnote{navigate\_to\_pose\_w\_replanning\_time.xml}, on distance traversed\footnote{navigate\_w\_replanning\_distance.xml}, proportional to robot speed\footnote{navigate\_w\_replanning\_speed.xml} or when the goal is updated\footnote{navigate\_w\_replanning\_only\_if\_goal\_is\_updated.xml}.
However, it is sometimes desirable for robotic applications to not dynamically replan (except when strictly required due to a potential future collision) to produce predictable behavior\footnote{navigate\_w\_replanning\_only\_if\_path\_becomes\_invalid.xml}.

Practical BTs typically include recovery behaviors to resolve the vast majority of errors that may occur during navigation due to dynamic scenes, uncertainty, or confined environments.
ROS~2 includes BTs that invoke context-specific actions in case of particular component failures.
They also contain a global recovery branch for handling system-level failures\footnote{navigate\_to\_pose\_w\_replanning\_and\_recovery.xml}. 
More advanced BTs include mixed-replanning\footnote{nav\_to\_pose\_with\_consistent\_replanning\_and\_if\_path\_becomes\_invalid.xml\label{footnote:fig}} and goal patience\footnote{navigate\_to\_pose\_w\_replanning\_goal\_patience\_and\_recovery.xml} to resolve an assortment of commonplace behavioral issues, such as oscillation due to frequent replanning and handling temporary blockages, respectively.

The navigate-through-poses configurations provide a similar behavior but contain hard pose constraints for via-points to navigate through on the way to a final goal.
These behavior trees make use of the `Compute Path Through Poses' planning node rather than `Compute Path To Pose'.
In addition, this class of BT removes waypoints passed so that subsequent replanning iterations contains only future looking constraints\footnote{navigate\_through\_poses\_w\_replanning\_and\_recovery.xml}.
These configurations may also contain different replanning mechanics, recoveries, and advanced features as described prior.

The BT tasks are not required to be represented by a goal or via-points - or even full autonomy.
Instead, a BT may be only interested in following a particular object\footnote{follow\_point.xml} with a collision free path or exploring new areas.
Other examples include algorithm assisted remote tele-operation, odometry calibration experiments\footnote{odometry\_calibration.xml}, and complete-coverage tasks.

\begin{figure}[ht]
    \centering
    \includegraphics[width=0.48\textwidth]{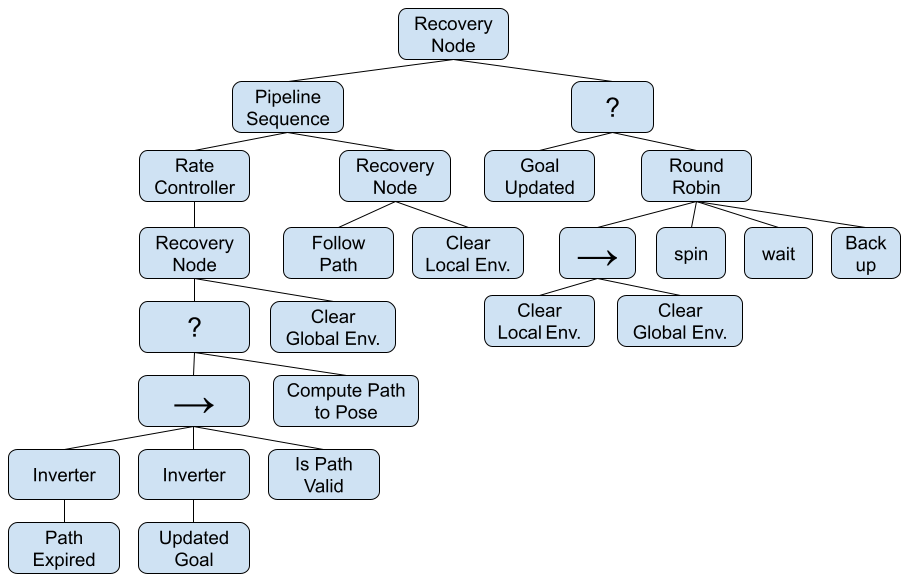}
    \caption{Mixed-replanning navigate-to-pose behavior tree \ref{footnote:fig}.}
    \label{fig:BT}
\end{figure}

As an illustrated example, Fig. \ref{fig:BT} shows the mixed-replanning navigate-to-pose behavior tree with recoveries described in footnote \ref{footnote:fig}.
The left branch from the root node contains its navigation logic, while the right is used for system-level recovery.
If a new goal is received during system-level recoveries, it will abort and reattempt navigation.
There are also context-specific recoveries clearing the global and local environmental models within the navigation branch when path planning or local trajectory planning fail, respectively.
In the navigation branch, mixed replanning is set to compute a new path if any of the following conditions are met: the path has expired after its duration of use, a new goal has been requested, or the path is found to be invalid due to sufficiently increased costs or collision.
Thus, this BT will immediately replan if the path or goal requires it, but also refine periodically (typically 5-30s).
This enables responsiveness to urgent needs while providing refinement as new information is processed.
Dynamic replanning is under a rate controller to limit the replanning rate slower than the tree's tick rate (100 Hz).

While not included in Fig. \ref{fig:BT}, it is advisable to leverage algorithm selector nodes as a recovery action to recompute a key function (path planning, trajectory planning) with another implementation.
When backup implementations and contextual recovery actions are well selected for an application, they can resolve most issues.
Further, the algorithm selectors allow a system to utilize many algorithms in their most useful contexts, rather than only a single method.

When designing behavior trees, it is prudent to create many independent nodes which are assembled in the tree's configuration for easy reusability and/or later customization.
It is also wise to place as much logic in the tree as possible, in contrast to client servers, to provide the maximum flexibility.
These principles were followed in the BTs outlined in this section, which has allowed them all to be composed from the same set of BT nodes from Table \ref{tab:behavior_trees}.

\section{Overview of State Estimation}
\label{sec:state_estimation}

For a mobile robot to successfully navigate through its environment, it needs to maintain an estimate of its state (pose, velocity, etc).
Such estimates are typically obtained through \textit{sensor fusion}, or the process of determining a robot's state relative to a reference frame given many noisy sensor measurements, such as IMUs, wheel encoders, pose measurements, GPS, and more.
The output of this fusion is an approximation of a robot's state variables and typically a measure of confidence regarding the estimate. 

ROS Enhancement Proposal (REP) 105 establishes the coordinate frames for common types of robotic systems to ensure interoperability of components \cite{rep105}. The transformation tree containing these coordinate frames is populated by the algorithms of this and the next sections and managed by the \textit{tf} library \cite{tf}.
\textit{tf} is utilized by a litany of applications to transform sensor data or relative vectors through time-varying reference frames - not simply for robot state tracking.

Within this formulation, the {\tt odom} to {\tt base\_link} transform can be generated by integrating a robot's wheel encoder data to provide a pose estimate that can be used for local control of the robot. That pose will drift over time, however, so REP 105 also specifies a {\tt map} to {\tt base\_link} transform\footnote{Technically, the {\tt tf} library does not permit a given coordinate frame having two parents, so we instead use the {\tt map} to {\tt base\_link} transform and the inverse of the {\tt odom} to {\tt base\_link} transform to generate a {\tt map} to {\tt odom} transform.} that represents a drift-free estimate of the robot's pose. This pose is useful for carrying out operations such as global planning. However, it is subject to discontinuities, making it a poor choice for control algorithms. The {\tt map} to {\tt base\_link} transform can be provided through a localization package, described in more detail in section \ref{sec:localization}.

Alternatively, both transforms can be provided through the use of state estimators. In the case of the {\tt odom} to {\tt base\_link} transform, this would typically mean fusing wheel encoder data with other continuous sources like angular velocity data from an IMU. For the {\tt map} to {\tt base\_link} transform, the pose estimate from the localization package might be fused with other global pose estimates or with the same sources fused in the {\tt odom} to {\tt base\_link} state estimator. This is useful if the localization package produces infrequent or irregularly-timed pose measurements, as it enables the global pose estimate to be generated at a fixed frequency.

While there are a myriad of algorithms for performing sensor fusion, two well-known and robust classes of approaches are Kalman filtering and factor graphs, both of which have implementations within the ROS~2 ecosystem.

\subsection{State Estimators}

\label{sec:rl}
\textbf{Robot Localization}
In the 60 years since its introduction, Kalman filtering has become one of the best-known and widely adopted approaches to state estimation \cite{kalman_filter}. The Kalman Filter algorithm manages the state estimate in two stages: predict and update. While the original Kalman Filter works for linear systems, the Extended Kalman Filter (EKF) addresses this shortcoming by linearizing the state transition function \textit{f} and measurement model \textit{h} around the current state. It then uses the resulting Jacobian matrices $\textbf{F}_t$ and $\textbf{H}_t$ in place of the original formulation's state transition matrix \textbf{F} and measurement model \textbf{H}, respectively in Eq. \ref{fh}.

\begin{equation}
\begin{split}
\label{fh}
\textbf{F}_t & = \frac{\partial f}{\partial \hat{\textbf{x}}_{t-1}} \\
\textbf{H}_t & = \frac{\partial h}{\partial \hat{\textbf{x}}_{t-1}}.
\end{split}
\end{equation}

A further extension known as the Unscented Kalman Filter (UKF) eschews the computation of Jacobian matrices in favor of using a set of \textit{sigma points} to carry out state projection \cite{unscented_filter}.
Each sigma point is effectively the most recent system state estimate $\hat{\textbf{x}}_{t-1}$, but with one dimension perturbed in proportion to the current covariance matrix $\hat{\textbf{P}}_{t-1}$.
Each sigma point is then projected through \textit{f} and the predicted state $\hat{\textbf{x}}_{t}$ and covariance $\hat{\textbf{P}}_{t}$ are recovered.

In ROS~2, the \textit{robot\_localization} (\textit{r\_l}) package provides implementations of both an EKF and UKF \cite{rl}. The package supports an unlimited number of sensor inputs from common ROS message formats such as {\tt Odometry}, {\tt Imu}, {\tt PoseWithCovarianceStamped}, and {\tt TwistWithCovarianceStamped}. Users also have the ability to control which dimensions from any given measurement source will be fused in the final state estimate. For example, if an IMU sensor produces faulty heading data due to magnetic interference from the magnetometer, that dimension can be excluded when fusing the IMU measurements. Pose data can also be fused \textit{differentially}, i.e., consecutive poses can be converted into velocities before being passed to the filter. This feature is useful when attempting to fuse world-referenced pose data from sources that do not agree with one another.

\textit{r\_l} uses an omnidirectional three-dimensional kinematic state transition function to estimate the 15-dimensional state vector given in Eq. \ref{rlstate}.

\begin{equation} \label{rlstate}
\textbf{x} = [x, y, z, \phi, \theta, \psi, \Dot{x}, \Dot{y}, \Dot{z}, \Dot{\phi}, \Dot{\theta}, \Dot{\psi}, \Ddot{x}, \Ddot{y}, \Ddot{z}]
\end{equation}

While the package does not support any other kinematic models, common models like the unicycle model can be obtained by clamping all three-dimensional state variables (as well as $\Dot{y}$ and $\Ddot{y}$) to 0. Furthermore, kinematic models such as Ackermann steering can be used if the measurement source provides absolute pose data and the appropriate process noise covariance matrix entries are increased to force the filter to trust measurements more than the filter.
In practice, this is sufficient for most mobile robotics applications and the most common method of sensor fusion for state estimation in ROS over the last half-decade.

The large number of parameters available to the user in \textit{r\_l} can be used to tune the filter's performance and tailor it to the user's application. In general, the following practices are recommended:
\begin{itemize}
\item If operating in a planar environment, set {\tt two\_d\_mode} to \textit{true}. While the filter will still continue to perform full 3D state estimation, dimensions outside SE2 will be set to 0.
\item For differential drive platforms, the only body-frame velocities that are typically relevant are $\Dot{x}$ and $\Dot{\theta}$. However, the 0 value for $\Dot{y}$ that is produced by the wheel encoder odometry should be fused in the filter. This prevents the filter from artificially generating non-zero $\Dot{y}$ in the state as the robot goes around turns.
\item Every linear and rotational dimension must have a reference (i.e., it must be measured by one of the user's sensor inputs). This measurement can be provided via absolute pose data \textit{or} velocity data. Prefer pose sources when fusing only a single input in that dimension. Prefer one pose and many velocity sources when fusing multiple sources for a dimension.
\item While the filter accepts linear acceleration measurements, they are generally insufficient for state estimation, as the double-integration of linear acceleration without any velocity or pose reference will lead to unbounded growth in that dimension.
\item Start with the minimum set required to provide a reference for every linear and angular dimension, and then add inputs one-at-a-time.
\item Take time to understand the coordinate frames of sensor data. Make sure that transforms are available for all data that is not provided in the principal coordinate frames.
\end{itemize}

\textbf{Fuse} The nonlinear state estimation problem can also be solved through iterative numerical optimization. It can be factored into the following least-squares form:

\begin{equation} \label{leastsquares}
\hat{\textbf{x}}_t = \arg \min_{\textbf{x}} \sum_i\left(\frac{\textbf{y}_i - g(\textbf{x})}{\Sigma^{\frac{1}{2}}}\right)^2
\end{equation}

where $\hat{\textbf{x}}_t$ is the optimal state estimate.
We obtain it by finding the state that minimizes the sum squared error between each observation $\textbf{y}_i$ and the transformed state $g(\textbf{x})$, weighted by the inverse square root of the covariance $\Sigma$.

Factor graphs represent this formulation graphically, as shown in Fig. \ref{fig:factorgraph}.
Two types of nodes are present in the graph: \textit{variable nodes} represent specific quantities that we want to estimate (e.g., the robot's pose at time \textit{t}) and \textit{factor nodes} represent constraints on those variables (e.g., kinematic constraints via a motion model).
In Fig. \ref{fig:factorgraph}, the variable nodes are the states $\textbf{x}_1$ through $\textbf{x}_4$, with the square factor nodes representing constraints between the connected states. This example shows a common use case for sensor fusion, in which asynchronous measurements arrive from different sensors. Each sensor measurement creates a constraint between two states, leading to two separate sets of vertices. In order to unify the graph, we create kinematic constraints between $\textbf{x}_1$ and $\textbf{x}_2$, $\textbf{x}_2$ and $\textbf{x}_3$, and $\textbf{x}_3$ and $\textbf{x}_4$. The graph is then optimized to find the most likely values for the states.

\begin{figure}[ht]
    \centering
    \includegraphics[width=0.48\textwidth]{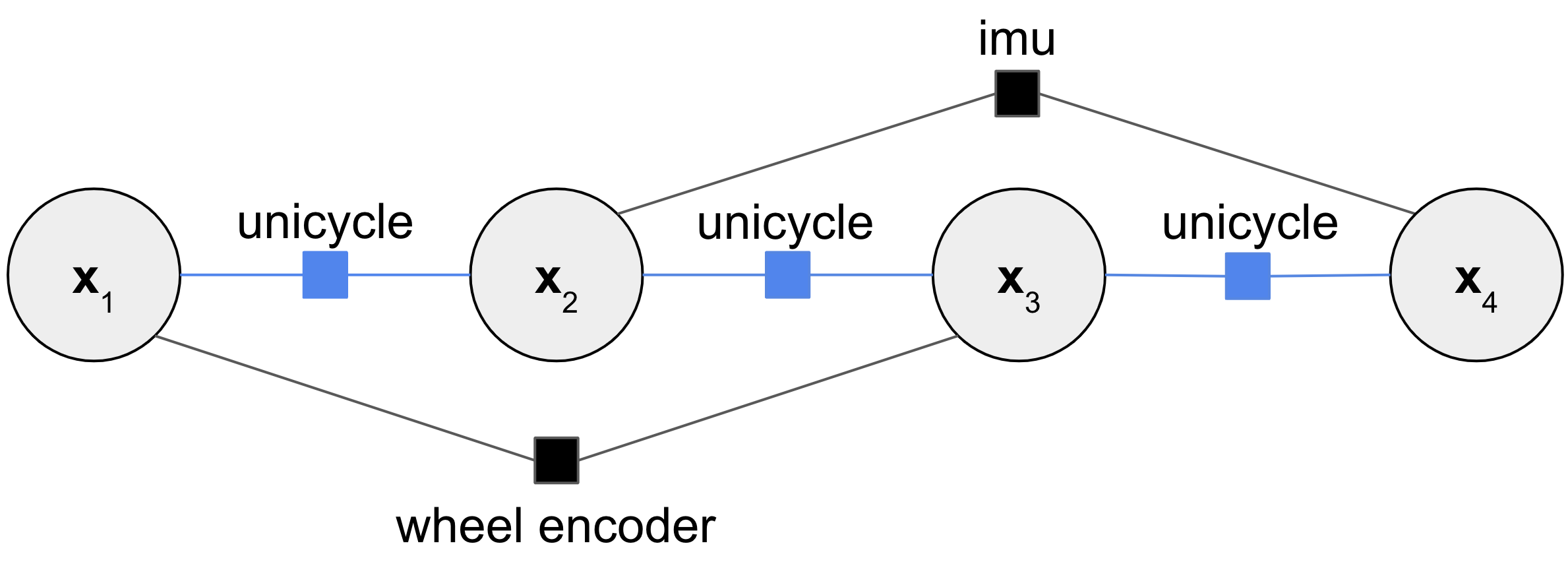}
    \caption{Example factor graph for a system with asynchronous measurements. A sensor measurement from a wheel encoder creates a constraint between state $\textbf{x}_1$ and $\textbf{x}_3$. A measurement from an IMU creates a constraint between state $\textbf{x}_2$ and $\textbf{x}_4$. The disjoint graphs are then connected via unicycle model kinematic constraints.}
    \label{fig:factorgraph}
\end{figure}

The ROS~2 \textit{fuse} package provides a factor graph framework that can be used in state estimation, mapping, calibration, and more \cite{fuse, fusegh}.
Within \textit{fuse}, each factor node can represent a measurement constraint or a kinematic constraint. Measurement constraints represent the error term in Eq.~\ref{leastsquares} as $\textbf{R}^{-\frac{1}{2}}\left(\textbf{z}_i - h(\textbf{x})\right)$, where \textit{h}, the sensor model, is a function that transforms the variables in the constraint into the space of the measurement.
The covariance matrix \textbf{R} controls how much weight the error term in question carries during the optimization process.
A small covariance would result in the measurement being heavily weighted.

Kinematic constraints represent cost in the form $\textbf{P}^{-\frac{1}{2}}\left(\textbf{x}_i - f(\textbf{x}_{t-1})\right)$, where $\textbf{x}_i$ is the current value of the variables connected to the constraint and the function \textit{f} is a state transition function that projects the variable values in $\textbf{x}_{t-1}$ forward in time.

\textit{fuse} makes extensive use of dynamically loaded plugins to manage the highly configurable graph construction and optimization process.
A number of plugins are already provided for common mobile robotics needs, but users may create and utilize their own plugins for custom requirements.
Each type of plugin are detailed below.

\textit{Variables}: Definitions of the quantities to be optimized. A variable may represent a single value, e.g., $x$ position, or it may represent more complex data, such as a timestamped two-dimensional pose. For example, the {\tt Orentation3DStamped} variable comprises a 3D orientation (represented as a quaternion) with an associated time stamp.

\textit{Constraints}: Connections for variables in the factor graph. Each constraint is responsible for generating some cost given the connected variables. These costs are then minimized during optimization. For example, the {\tt RelativePose2DStampedConstraint} produces a cost for two poses in the graph that is proportional to the difference between those poses and the measured difference as provided by a sensor measurement.

\textit{Sensor Models}: Models of variables with associated constraints generated from received sensor data. The variables and constraints are then passed to the optimizer. For example, the {\tt Imu2D} sensor model produces the relevant two-dimensional orientation and angular velocity data variables to add to the graph, as well as constraints on those variables as given by the measurement itself.

\textit{Motion Models}: Models of kinematic constraints between timestamped variables. These are generated on request and passed to the optimizer. For example, the {\tt Unicycle2D} kinematic model generates a constraint between two pose variables in the graph. If those poses are not feasible given a unicycle kinematic model, the constraint will generate non-zero cost.

\textit{Optimizers}: Build and optimize the graph. The core nodes in \textit{fuse} are effectively wrappers around optimizer plugins. For example, the primary state estimation node in \textit{fuse} is known as the \textit{fixed-lag smoother}. It wraps an optimizer of the same name.

\textit{Publishers}: Extract relevant quantities from the optimizer and publish the ROS~2 messages containing results. For example, the {\tt Pose2DPublisher} publishes the most recent optimized pose and covariance from the graph as both a { \tt geometry\_msgs/PoseStamped} ROS~2 message as well as a transform to be consumed by other nodes via \textit{tf}.

Each iteration, sensor model plugins will receive sensor data and create constraints based on them.
The optimizer will receive these constraints and request the motion model plugin to generate constraints to connect the sensor model constraints to the existing graph.
These new variables and constraints are optimized over via the optimizer to compute optimal values.
Finally, the publisher will return the results over ROS~2 topics to interested nodes.

\textit{fuse} has a number of advantages over \textit{r\_l}, both due to software design and affordances provided by factor graphs:

\begin{itemize}
  \item \textit{fuse} is more flexible and extensible. In addition to \textit{r\_l}'s support of an arbitrary number of sensors, \textit{fuse}'s plugin-based architecture allows configuration of the state vector, motion, and sensor models.
  \item Relative pose measurements are better supported in \textit{fuse}.
  \item Linearization errors are reduced in \textit{fuse} at every iteration in the optimization.
  \item \textit{fuse} can be used for applications outside sensor fusion, mapping, and calibration; one such application is robot control.
\end{itemize}

However, these features come at an increased compute cost. The experiments in section \ref{sec:comparison} suggest that \textit{fuse} could require as much as 4x the CPU of \textit{r\_l}. As \textit{r\_l} is fairly lightweight, this increase is generally acceptable for the vast majority of applications.

For those particularly resource-constrained applications, there are a number of parameters that may be tuned to improve compute time.
Reducing the length of the smoothing window of the fixed-lag smoother and the frequency of update can improve the performance. 
Further, \textit{fuse} uses Google Ceres' automatic differentiation in many of the plugins to compute Jacobians \cite{ceres}.
While this is convenient, it is not as efficient as analytically derived Jacobians, which Ceres also supports.

\subsection{Comparison}
\label{sec:comparison}

To compare the two state estimation systems in ROS~2, an experiment was designed and run on a 3.8 GHz AMD Ryzen 5800x CPU.
A mobile robot equipped with wheel encoders, an IMU, and a laser scanner was driven along a 541-meter long route through a commercial environment to collect a dataset, ending within 10 cm of the starting pose.
The odometry and IMU data was collected at 25 Hz.
\textit{r\_l}'s EKF node and \textit{fuse}'s fixed-lag smoother node each fused linear and angular velocities from the wheel encoders and angular velocities from the IMU.
ROS~2's \textit{amcl} package provides the ground-truth information (Sec. \ref{sec:amcl}). The fixed lag smoothing window in \textit{fuse} was set to 0.5 seconds and the covariance was extracted from the graph at 1 Hz.

The results can be seen in Fig. \ref{fig:se_comp} and Table \ref{tab:se_comp}.
At the end of the path, the \textit{fuse} estimate was 1.44 meters closer to the ground truth position than the \textit{r\_l} estimate, though both estimates were well below 5.4 meters, or 1\%, of the ground-truth pose.
The CPU usage for \textit{fuse} was approximately 3.7x that of \textit{r\_l}, with a slightly larger standard deviation.

\begin{table*}[ht]
 \caption{State Estimators Comparison}
 \label{tab:se_comp}
 \begin{center}
 \begin{tabular}{ |c|c|c|c|c|c| }
   \hline
   State Estimator & Update Frequency (Hz) & Total Distance (m) & Total Error (m) & Error (\% Distance Traveled) & CPU \% \\
   \hline
   \hline
   robot\_localization & 30 & 542.16 & 4.25 & 0.78 & 1.38 $\pm$ 1.17 \\
   \hline
   fuse & 20 & 542.46 & 2.81 & 0.52 & 5.19 $\pm$ 1.37 \\
   \hline
 \end{tabular}
 \end{center}
\end{table*}

\begin{figure}[t]
    \centering
    \includegraphics[width=0.48\textwidth]{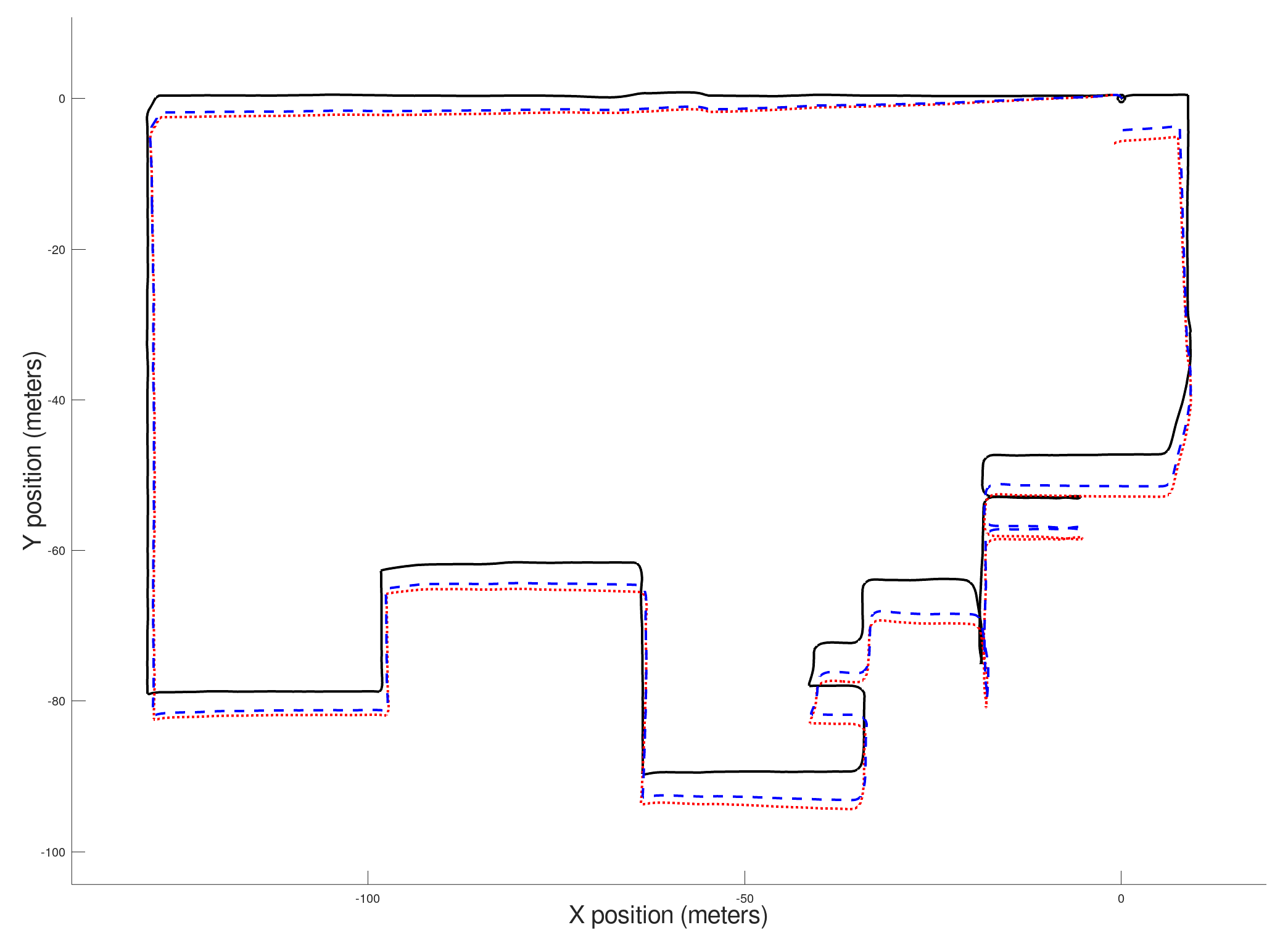}
    \caption{Three state estimates of the same circuitous path. Black path: \textit{amcl}-generated ground truth path; dotted red path: estimate of the path fusing only wheel encoder and IMU data for \textit{r\_l}; dashed blue path: estimate of the path using the same data, but fused via the \textit{fuse} package.}
    \label{fig:se_comp}
\end{figure}

\section{Overview of Localization and Mapping}
\label{sec:localization}

Localization is the process by which the robot determines where it is with respect to a global frame of reference.
As opposed to state estimation, this typically occurs at a reduced rate and provides corrections to odometric drift.
A localization system attempts to match its sensor data to a known model of the environment or build the model simultaneously, e.g., Simultaneous Localization and Mapping (SLAM).
Commonly, localization is preformed using 2D laser scanners, 3D lidars, or cameras - though techniques exist using more exotic and/or mixtures of sensors.
Localization's output is a pose and confidence relative to a world model.
This pose solves the REP 105 portion of the \textit{tf} transformation tree from {\tt map} to {\tt odom}.

In ROS~2, the standard localization and SLAM implementations continue to be 2D laser scanner techniques due to their robustness in dynamic and changing environments over long durations.
While strides have been made in visual techniques, the field is still rapidly evolving and even relatively mature methods like ORB-SLAM3 are insufficiently robust for large classes of practical robotics tasks \cite{vslam}.
They are generally sufficient for a number of small-scale or short-duration applications, but official ROS~2 support requires further substantive improvements. 

\subsection{Localization}
\label{sec:secloc}
\label{sec:amcl}

\textbf{AMCL} One popular approach for solving the localization problem is through the use of particle filters \cite{particle_filter}, which enable robust state estimation in dynamic environments and in the presence of noisy sensor data. Particle filters represent non-Gaussian statistical distributions using a discrete set of hypotheses (particles). In the case of mobile robot localization, the distribution represents the probable pose of the robot. 

If an initial pose is not set, a particle filter may initialize the particle cloud by uniformly distributing it throughout the space.
Otherwise, the particles are distributed around the provided start pose of the robot.
At each update step, the particles are projected using a kinematic model of the robot with noise perturbations.
The particles are then weighted based on the likelihood of their sensor observations at that pose.
The filter may then resample the particles, that is, generate a new particle distribution by selecting particles from the existing distribution.

The \textit{amcl} (Adaptive Monte Carlo Localization) package contains a pure and fully-parameterized implementation of the grid particle filter localizer, Fig. \ref{fig:amcl} \cite{prob_rob}.
The package exposes numerous parameters for controlling the full behavior of the filter: minimum and maximum particle counts, resampling periods, statistical weights, motion model noises, and more.
State projection is handled via a motion model that supports differential drive or holonomic bases with configurable characteristics.
Particle weight assignment is performed by taking the current laser scan of the environment in the sensor frame, transforming it based on the pose of each particle, and then attempting to match the scan to a rastered map of the environment.
Particles that have better matches with the map will therefore have higher weights, and will be more likely to be selected in the next resampling step.

Like the odometry model, the laser model is highly configurable.
Two main models are supported: the \textit{beam model} performs ray tracing from each particle to generate a model laser beam, and then compares the ideal beam with the sensed laser measurements \cite{beammodel}.
The beam model models four sources of error due to physical phenomenon with the laser scanner and environment: sensor measurement noise, changes in the environment, failures to measure, and random noise.  
The \textit{likelihood field model} diffuses the map via a Gaussian distribution into a lookup grid such that each cell represents the range from that cell to the nearest occupied cell in the original map \cite{lhfieldmodel}.
Each laser scan point is then projected onto the lookup grid, and the score for the scan is proportional to the sum of the values of the cells into which the projected points fall.
This models sensor noise using the Gaussian distribution, but also explicitly models errors from sensor failures and random measurements in the same way as the beam model. 
This can be more optimal computationally as well as resolve lack of smoothness concerns with the beam model due to many small, dispersed occupancy grid entries in complex settings.
The performance characteristics of the \textit{amcl} package is well researched and can achieve localization accuracies of 5 cm or better in many practical environments \cite{amcl_acc}.

\begin{figure}[ht]
    \centering
    \includegraphics[width=0.40\textwidth]{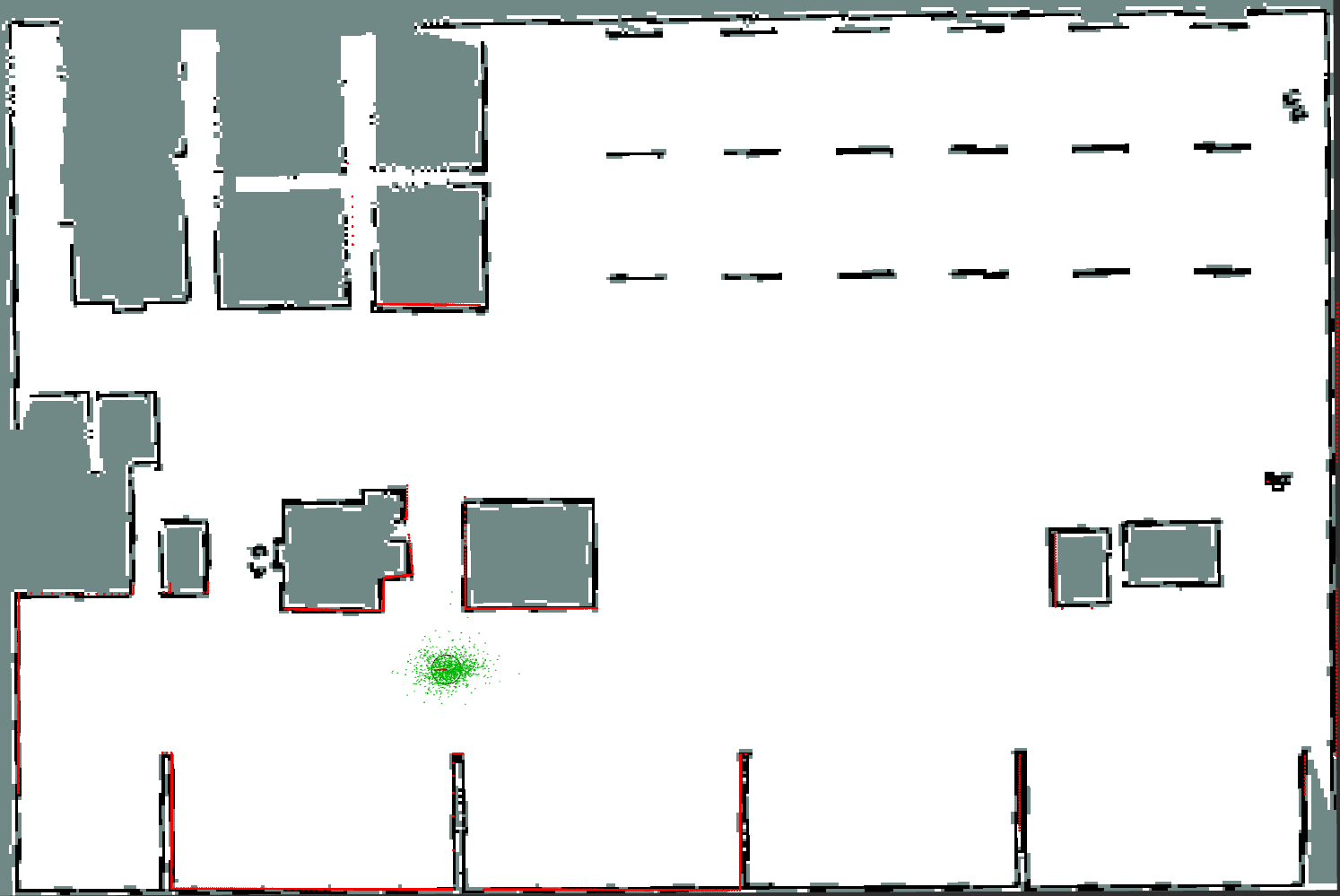}
    \caption{AMCL's particle cloud localizer in a warehouse environment with a laser scanner. Map in black, laser scan readings in red, particles in green, and the robot's pose as the frame with black circular footprint.}
    \label{fig:amcl}
\end{figure}

\textbf{GPS Localization} Localizing a robot using GPS sensors is a popular approach for outdoor robot platforms. While subject to drift in some environments, GPS data provides a globally unambiguous positional reference and reduces the need for techniques that rely on expensive sensors. The key to integrating GPS data in a robot's state estimate is the computation of the transform between the robot's world frame (e.g., the {\tt map} frame as specified in REP-105) and the GPS frame, most commonly the WGS-84 geodetic datum \cite{wgs84}.

Generating this transform requires knowledge of the robot's pose in a world frame, the robot's latitude and longitude as provided from a GPS sensor, and the robot's heading in an earth-referenced frame, as might be obtained via an IMU with a magnetometer.

The \textit{robot\_localization} package described in Section \ref{sec:rl} provides the NavSat Transform Node, whose purpose is to calculate this transformation. It does this by converting the GPS position and earth-referenced heading into a position in the Universal Transverse Mercator (UTM) coordinate frame \cite{utm}. From there, the transform to the robot's world frame is trivial. The output of NavSat Transform Node is a pose that can be fused into the robot's state estimate or directly used for localization with respect to the robot's world frame.
The accuracy of GPS localization is limited by the environment, quality of the GPS receiver, and whether RTK (Real-time kinematic) GPS corrections are available in the area.
When RTK is available, accuracies can reach 2 cm or better \cite{rtk}.

\subsection{Mapping}

Mapping solves the fundamental problem of how a robot understands its environment.
It uses sensor data and potentially robot state estimates to generate a globally accurate model of the world.
This world representation is used for many tasks, including global planning and localization, when not using GPS or other global positioning systems.
Succinctly, mapping is the process of learning a world's contents from sensor data.
For mapping techniques in ROS~2, the map is represented as an occupancy grid, or a grid map of cells representing the probability of occupancy.
They also perform mapping and localization simultaneously, e.g. SLAM. 

\textbf{Cartographer} provides real-time simultaneous localization and mapping in both 2D and 3D for multiple platforms and sensor configurations using pose-graph optimization \cite{cartographer}.
Laser scans are added into submaps over short periods of time to create locally accurate maps with current measurements.
Scan matching is performed on these local submaps and the scan is inserted into the submap at its best estimated pose.
When a submap is completed, it takes part in a loop closure process which compares submaps and local scans to refine the global occupancy grid model.

While Cartographer is a good option for some users, it is worth noting that out-of-the-box results with Cartographer are typically poor.
It is possible to achieve 3-5 cm accuracy using Cartographer, but requires extensive professional tuning using hardware platforms with high-quality odometry.
Further, the original author's organization has halted development and support.
ROS~2 support is managed by a community effort but has not received significant attention due to the complexity of the codebase.
Thus, it is difficult to recommend the use of Cartographer. 

\begin{figure}[ht]
    \centering
    \includegraphics[angle=90,width=0.48\textwidth]{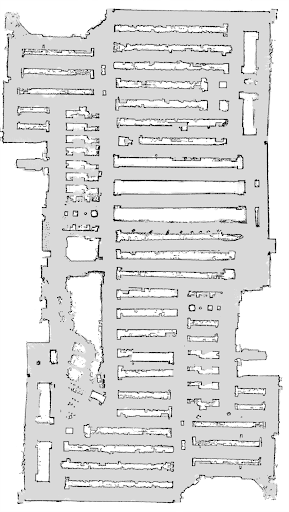}
    \caption{Map generated by SLAM Toolbox \cite{slam2}.}
    \label{fig:slam}
\end{figure}

\textbf{SLAM Toolbox} is a set of tools and capabilities for 2D SLAM built upon the OpenKarto SLAM library \cite{slam}.
SLAM Toolbox is also a pose-graph optimization method.
It matches incoming sensor readings with a recent rolling buffer of scans to provide a corrected pose.
This uses a correlation grid constructed from the buffer of local measurements to match the incoming scan at a coarse half-resolution, followed up by full resolution for fine tuning the results.

The incoming scan with the initially corrected pose is then evaluated for loop closure.
Candidate scans are found which are near the incoming measurement but not connected locally in the graph.
The candidates are matched against the scan at a coarse resolution.
If the match is strong enough, a full resolution match is performed on the fine correlation grid.
If again the match is sufficiently strong, a constraint is added; then the pose-graph optimization process is performed. 

Changes were made to the OpenKarto library to increase speed of scan matching using multi-threading and replace the Sparse Bundle Adjustment with Google Ceres to provide more flexible optimization settings. It can reliably map spaces greater than 100,000 sq ft in real-time (fig. \ref{fig:slam}), enables serialization of maps for continued mapping, allows for manual manipulation of the pose-graph, and comes with several operating modes.

\section{Overview of Key Utilities}
\label{sec:utils}

This section provides overviews of new and unique utilities in ROS~2. 
They contribute intuitive user interactions and create safer robotic systems by commanding, regulating, and validating the autonomous navigation system or its outputs.

\textbf{Lifecycle Manager} governs the state transitions of the nodes. It is used to orchestrate the bringup and shutdown of the system safely and deterministically to guarantee all constituent components of ROS~2 are active and ready before the first command is processed. It also maintains a connection to each server throughout execution to deactivate the system should any component fail. It may also re-activate the system after an error is handled.  

\textbf{Collision Monitor} checks for breaches in zones around the robot from sensors such as pointclouds, laser scans, or sonars to prevent imminent collisions, bypassing navigation's planners in case of sudden and urgent changes. These zones may ask a robot to slow to a reduced speed or immediately stop when a sufficient number of returns are detected. 
These zones may also be scheduled based on velocity.
The monitor also has an \textit{Approach} mode to project a robot's footprint forward in time and reduce the speed to assure the system is always a certain duration away from potential collisions. An example can be seen in Fig. \ref{fig:polygons}, showing the three primitive types of collision monitoring supported: stop, slowdown, and approach polygons. Note that any convex or concave polygon may be specified in the primary robot body-fixed frame.

\begin{figure}[ht]
    \centering
    \includegraphics[width=0.35\textwidth]{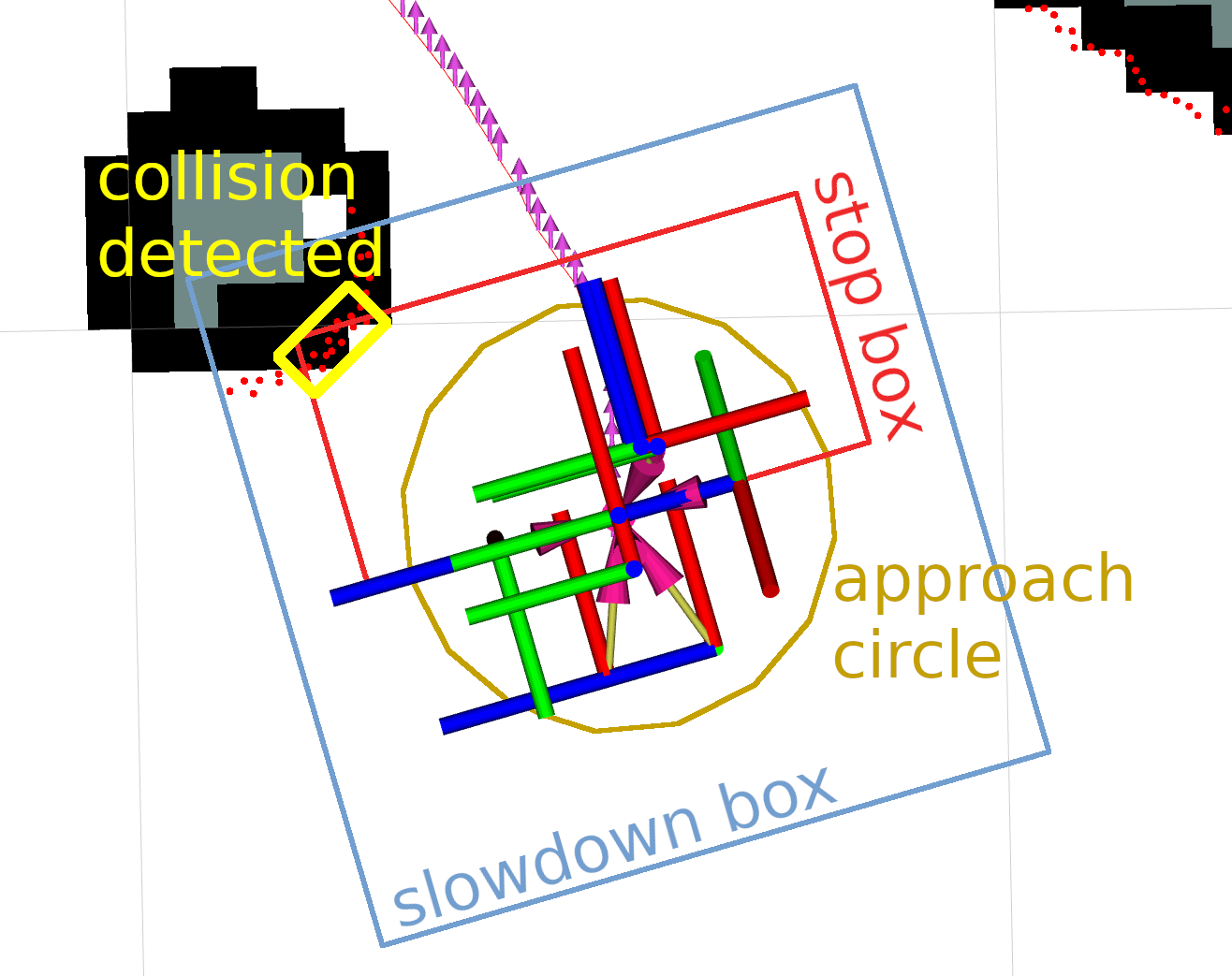}
    \caption{Collision Monitor zones on a sample robot.}
    \label{fig:polygons}
\end{figure}

\textbf{Velocity Smoother} smooths the output velocities from the navigation system by applying dynamic constraints based on the current robot state. This improves smoothness of the commands sent to a base controller, which can increase the longevity of the hardware. When set to a higher rate than incoming velocity commands, it will also interpolate the commands by dynamic limits providing even smoother output to the hardware at a regular interval.

\textbf{Simple Commander} is a Python3 API to command Nav2 at all levels from the BT Navigator down to its constituent servers allowing a user to have a `pythonic' user experience. It may be used to create autonomy systems leveraging Nav2 (or bypassing BT Navigator to define custom hand-engineered logic) in Python3.
It contains a cost map API for handling incoming occupancy grid messages and querying it for information. 

\textbf{Waypoint Follower} is a server used to request navigation to many waypoints and performing some action at each. The waypoint follower uses task executor plugins to complete the action at each waypoint (ex. wait for user input, take a photo, pick up an object, etc) which are dynamically loaded at run-time and fully configurable. The server also supports looping a series of waypoints for repetitive actions. 

\textbf{Command and Control Panel} is an Rviz user interface for interacting with Nav2 for testing, evaluation, or getting started. It provides tools to command the robot with navigate to pose, navigate through poses, and waypoint following tasks. It also visualizes the action feedback, can cancel ongoing tasks, and control the system's lifecycle states.

\textbf{Robot simulation} eases the process of development, testing, and reproduction.
Nav2 provides out-of-the-box support for the Gazebo simulator, a sister project of ROS, for testing and evaluation before deploying new capabilities onto hardware robots \cite{gazebo}.
Gazebo has been developed and used for over a decade by ROS Community and contains numerous features helpful for building practical robot systems, including:
\begin{itemize}
  \item Wide variety of supported sensors
  \item Plugin-based physics with rigid body dynamics
  \item A client-server architecture allowing distributed computation
  \item Fully integrated with ROS
  \item Cross-platform support with cloud integration
  \item XML described simulation environments using Simulation Description Files (SDF)
\end{itemize}

Nav2 provides a set of simulations containing a robot platform in various environments including a sandbox and a warehouse environment for easy testing and evaluation of the framework. As an alternative, additional environments can be easily created for a robot platform, as depicted on Fig. \ref{fig:GazeboTB3}.

\begin{figure}[ht]
    \centering
    \includegraphics[width=0.48\textwidth]{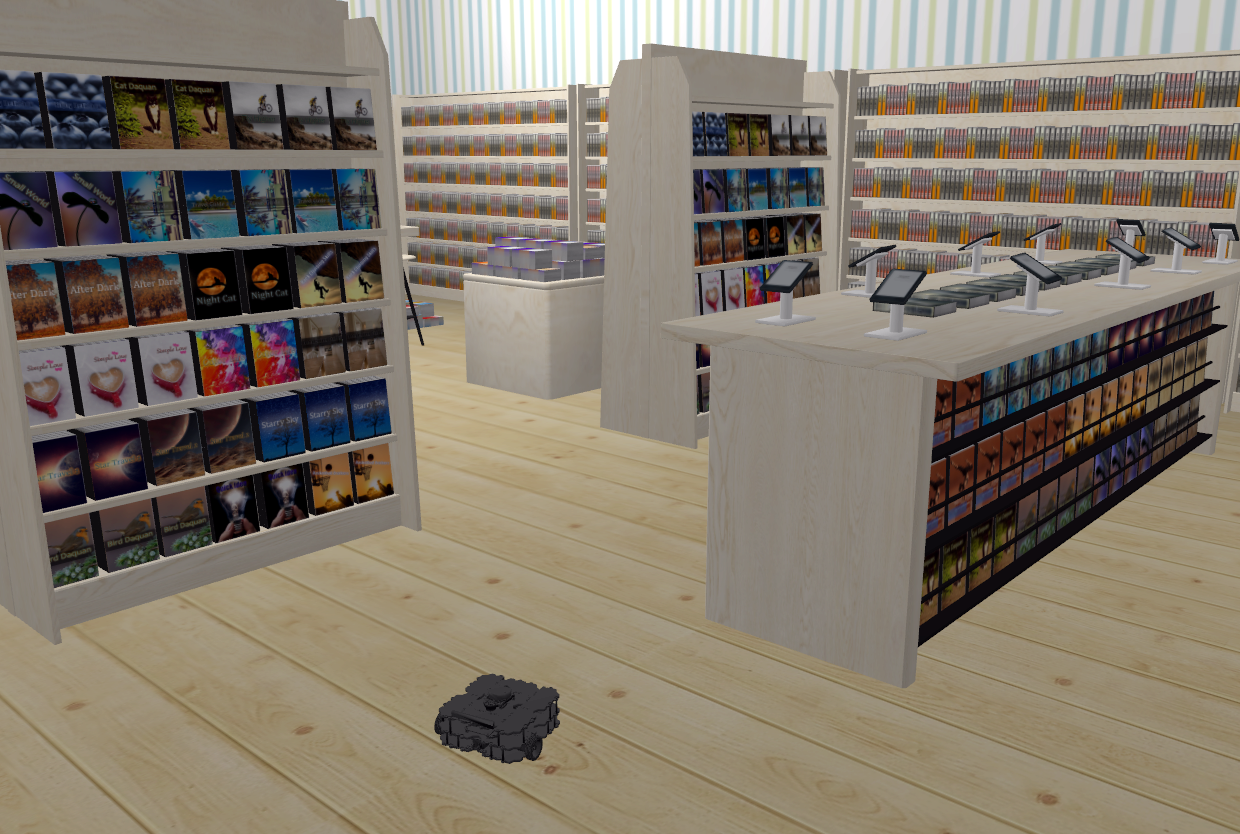}
    \caption{Example of TurtleBot3 Waffle robot, operating in the AWS Bookstore environment simulated by Gazebo \cite{aws_bookstore}.}
    \label{fig:GazeboTB3}
\end{figure}

The Gazebo brand has two distinct products within it: Gazebo (aka Gazebo Classic) and Ignition-Gazebo (aka Ignition). 
Ignition-Gazebo is a re-write of the legacy software to provide a modular, plugin-based simulation framework. 
Although not yet as fully featured, Nav2 will soon provide support for Ignition as well \cite{classic_ignition_comparison}.

\section{The Future of ROS~2 Mobile Robotics}
\label{sec:future}

The previous sections discussed the algorithms, behaviors, and features currently available in Nav2, and ROS~2 in general. 
These have been the product of several years of dedicated collaborations between industry, academia, and individual contributors across the globe. 
These efforts continue today to develop additional missing technologies for robotics research and commercialization. 
This section highlights a selection of notable ongoing and future-looking projects in the ROS~2 mobile robotics ecosystem being pursued.

There are many situations where it is necessary to plan global paths based on a reduced representation of the environment, rather than in free-space where the robot may travel anywhere.
Large-scale warehouses may be so massive that it is unrealistic to replan entire routes within a reasonable time-frame using free-space planning.
They may also require the robot to travel on pre-assigned routes wherever possible.
Further, local trajectory planning in non-planar environments (e.g. parking garages, outdoor unstructured environments) may require the use of traversibility estimation in the local time-horizon.
Rather than creating and storing similar height-maps for global planning, routing is frequently utilized.
In development is a Route Server analogous to the Planner Server to enable navigation-graph global route planning to handle all of these situations within the Nav2 framework.
It will use a pre-defined graph to plan through spaces with configurable constraints and directionality.
The graph may represent a hand-selected set of points, automatically generated nodes and edges from a visibility graph, the result of a Probabilistic Roadmap generator, and more.
This will unlock the ability for directional and non-directional lanes or bounded navigation regions, coarse global routing through massive spaces which may be refined within a window, algorithmically-defined reduced free-space planning, and general outdoor navigation.



Additionally, work is being conducted to extend Nav2's provided environment models out of two-dimensions.
Extensions are being designed to facilitate height maps with traversibility estimation for outdoor, urban, and natural environments.
This will enable new classes of robot systems to leverage Nav2 natively without coarsely approximating terrain information using cost or risk maps.
It also improves traditional indoor robot systems, facilitating more accurate ground information to avoid small objects and exploit ramps.

Finally, in the future, more attention will be focused on the problems of two- and three-dimensional localization.
AMCL, used commonly in ROS, has a number of shortcomings and the state of the art in localization has improved since its creation.
A design has been proposed to create a localization framework for building both 2D and 3D localization systems atop of modular components.
In the future, this will allow developers to swap out individual components of standard implementations to build use case-specific or novel localization systems without being required to recreate all other components.
Its initial release will contain an analog to AMCL, but also recent algorithms such as DT-NDL-MCL and pose-graph based solutions \cite{ndt}.

\section{Conclusion}
\label{sec:conclusion}

This paper surveys the state-of-the-art in ROS~2 navigation, as of November 2022, with key metrics and expert insights from principal ROS developers and maintainers. 
The redesign of Nav2 has triggered a renewed interest in mobile robotics technology in the ROS ecosystem.
Since 2018, Nav2 and related projects have continued to develop modern algorithms to augment historically capable methods in continual use since Willow Garage.
New global path planning, local trajectory planning, cost map, and state estimation algorithms have been introduced to enhance navigation capabilities to support a broader set of vehicles, environments, and applications.
Moreover, entirely new classes of algorithms and systems have been added to hone performance: such as path smoothers, behavior trees, and numerous utilities.
This innovation has not terminated - new developments come on a quarterly basis and continues to thrive amongst an ever increasing number of commercial, academic, and individual contributors.

\bibliographystyle{IEEEtran}
\bibliography{references}

\appendices

\section{Algorithm Selection by Robot Model}
\label{algo_selection}

Sections \ref{sec:planners} - \ref{sec:smooth} compare global path planners, trajectory planners, and smoothers between each other with brief descriptions of their most useful robot model types in Tables Tables \ref{tab:planners} - \ref{tab:smoothers}.
However, it is useful to provide an expanded discussion on these summaries and provide algorithm optimal use cases, which is further summarized in the table \ref{tab:algorithms}.

\begin{table*}[!ht]
  \caption{Summarized Algorithm Comparison}
  \label{tab:algorithms}
  \begin{center}
  \begin{tabular}{ |p{3cm}|p{13.8cm}| }
    \hline
    \multicolumn{2}{|c|}{\textbf{Planners}} \\
    \hline
    NavFn & Good general purpose and performant holonomic planner \\
    \hline
    Theta* & Best at operating in environments with long, straight, and potentially non-axially aligned corridors or open spaces \\
    \hline
    Smac 2D-A* & Best on circular robots in a heterogeneous fleet containing other non-circular or Ackermann robots for consistent behavior \\
    \hline
    Smac Hybrid-A* & Best for car-like or Ackermann robots \\
    \hline
    Smac State Lattice & Best for for large non-circular non-Ackermann robots or when requiring a finite set of describable robot motion primitives \\
    \hline
    \hline
    \multicolumn{2}{|c|}{\textbf{Controllers}} \\
    \hline
    DWB & Decent reactive choice when operating in static scenes or far from obstacles, with fully replacable objective functions \\ 
    \hline
    MPPI & Best predictive choice for dynamic scenes and confined areas for all platforms, with fully replacable objective functions \\
    \hline
    TEB & Historical predictive choice for dynamic scenes and confined areas for all platforms, with moderately tunable optimization \\ 
    \hline
    RPP & Geometric velocity-regulated controller for non-holonomic robots which must follow a path without deviation \\
    \hline
    Graceful & Good spiral technique for differential-drive robots when movements must be smooth without dense paths \\
    \hline
    Rotation Shim & Rotation primitive before applying other algorithms to rotate toward path heading, if possible, for smooth path tracking \\
    \hline
    \hline
    \multicolumn{2}{|c|}{\textbf{Smoothers}} \\
    \hline
    Simple Smoother & Good for smoothing holonomic paths to remove defects caused by oscillation and widing out turning maneuvers \\
    \hline
    Constrained Smoother & Best for feasible paths to guarantee drivability optimizing for cost, curvature, and smoothness -- computationally expensive  \\
    \hline
    Savitzky-Golay Smoother & Good for removing localized defects caused by gradient descent, cell binning, or instantaneous changes (NavFn, Theta*) \\
    \hline 
  \end{tabular}
  \end{center}
\end{table*}

\subsection{Circular Differential-Drive}
\textbf{Recommendation: NavFn + MPPI}

Circular differential-drive robots typically leverage infeasible global path planners. 
NavFn is the recommended algorithm for such cases, as a high performance and highly deployed implementation used for over a decade. 
If deploying a circular differential-drive robot in injunction with other non-circular platforms, the Smac 2D-A* may be a good option to have behavioral consistency with other robots using the feasible Smac Planners - in exchange for increased compute time. 
When using a robot in non-axially aligned maps, Theta* can be most useful to generate straight-line paths at arbitrary angles.

All of Nav2's trajectory planners are also built with optimizations for circular robots and will work well for this class of robot, but with different behavioral attributes.
The use of MPPI or TEB may be beneficial if in highly dynamic settings or where needing to get out of tight spots.
These predictive controllers, however, come at an elevated computational cost. 
Reg. Pure Pursuit is useful for exact-path tracking behavior in static scenes or where deviation from paths is not permissible while having an exceptionally small computational footprint.
DWB is a middle-ground all-around option when not in dynamic scenes but deviations from the path would be beneficial - for instance to optimize for smoothness or increased distance from static obstacles than the global infeasible plan provided. 

Typically, path smoothers for circular differential drive robots are not necessary for good performance.
However, infeasible global plans may be improved with the use of the Savitzky-Golay smoother to remove localized path defects as the result of gradient descent or cell binning, but these changes are largely superficial and do not manifestly impact system behavior when paired with a Nav2-provided trajectory planner.

\subsection{Non-Circular Differential-Drive}
\textbf{Recommendation: Smac State Lattice + MPPI + Simple Smoother.}

Non-circular differential drive robots generally require the use of feasible planners to guarantee drivability in confined spaces whereas they may not rotate in place without collision.
The use of the Smac State Lattice planner is recommended as a high performance option with adjustable motion primitives (including rotating in place). 
However, Smac Hybrid-A* could also be appropriate if a minimum control set is difficult to model. 
If non-circular robots are in settings where they may generally rotate in place anywhere they might travel (e.g. not confined or large), infeasible planners like the NavFn global planner may be deployed instead.

All trajectory planners are built to work with non-circular robots as well, using the robot's footprint to collision check. 
The same advice applies as described above for circular robots, but with an added computational cost for these heavier collision checking requirements. 
However, from experience with non-circular robots, it is recommended to leverage the Rotation Shim controller to rotate the platform to the relative path heading before path tracking when using an infeasible planner with either DWB or Graceful.
While not required, it is a step-up in an observer's predictability as these may struggle to make massive angular changes effectively. 

Providing large, non-circular robots with a smoother global path can impact small oscillatory behaviors notable on non-circular robot platforms due to their shape. 
Due to this, the simple smoother is recommended as a fast and reasonably effective solution to loosen turns and straighten out oscillations. 
The constrained smoother may also be applied, but it is computationally intensive and considers more criteria than may be required for a typical differential-drive robot application.

\subsection{Omni-Directional}

Omni-directional robots may follow the same advice outlined in the differential-drive discussions above following the appropriate circular vs non-circular distinction.
The only notable difference is that Reg. Pure Puruit behavior is not possible for omnidirectional robots to fully utilize its drive-train's lateral movement capabilities. While it may be employed, it will act as though it is a differential drive robot. 
All other trajectory planners may utilize the full capabilities of omnidirectional platforms.

\subsection{Ackermann}
\textbf{Recommendations:}
\begin{itemize}
    \item \textbf{Smac Hybrid-A* + MPPI,}
    \item \textbf{Smac Hybrid-A* + RPP + Constrained Smoother}
\end{itemize}

Ackermann robots should typically use the Smac Hybrid-A* planner, a purpose-built planner for Ackermann or car-like robots.
However, the State Lattice planner is also applicable if an application demands a finite number of totally predictable maneuvers or behaviors.

It is recommended to use TEB, MPPI or RPP for Ackermann robots.
TEB and MPPI provide predictive controllers while RPP provides exact path following behaviors to car-like robots. 
While DWB may be applied to Ackermann robots, the trajectory generator for curvature constraints has not yet been contributed to the project. 

The constrained smoother is recommended to be paired with the global plan with Ackermann robots to improve drivability and proximity to obstacles.
However, this comes at a non-trivial computational overhead and it may only be required for RPP, as MPPI and TEB will internally minimize similar constraints as part of their formulations.
The selection of smoother should be made in conjunction with the selection of path tracking behavior (trajectory planner).

\section{Smac Planner Framework}
\label{smac_appendix}

The \textit{Smac Planner} within Nav2 is a search-based planning framework with multiple algorithms including Hybrid-A*, State Lattice, and 2D-A*.
It contains a shared and highly optimized A* algorithm for which each algorithm utilizes templated by \textit{Node Types}.
These node types contain both the node's search information (e.g. visited, queued, cost) but also contains the planner's search characteristics. 
Specifically, they contain the methods specific to each planner's behaviors such as heuristics, expansion characteristics, and traversal cost functions. 
All of these functions are shared via static members to gain the performance equivalent of a standalone application. 
This framework minimizes boilerplate such that new planners may be added in as few as 50 lines of code (with an average of approximately 200).


\end{document}